\documentclass[runningheads]{llncs}
\usepackage{graphicx}
\usepackage{comment}
\usepackage{amsmath,amssymb, bm} 
\usepackage{color}

\usepackage{booktabs} 

\usepackage{enumitem}
\usepackage{dsfont}

\usepackage{nicefrac}       
\usepackage{microtype}      
\usepackage{gensymb}
\usepackage{xcolor}
\usepackage{multirow}
\usepackage{graphicx}
\usepackage{tabularx}
\usepackage{setspace}

\usepackage{makecell} 
\usepackage{pdfpages}
\usepackage{adjustbox}
\usepackage{balance} %
\usepackage{subcaption}
\usepackage{array} 

\usepackage[linesnumbered,ruled]{algorithm2e}
\usepackage{algpseudocode}
\usepackage{mathtools}
\usepackage{cuted}
\usepackage{afterpage}
\usepackage{bbm}
\usepackage{xspace}

\newcommand{\ignorethis}[1]{}

\DeclareMathOperator*{\argmin}{argmin}

\SetCommentSty{mycommfont}

\newcommand{\tabcell}[2][c]{\begin{tabular}[#1]{@{}c@{}}#2\end{tabular}}

\usepackage{amsmath,amsfonts,bm}

\def\eqref#1{equation~\ref{#1}}

\def\1{\bm{1}}

\newcommand{\test}{\mathcal{D_{\mathrm{test}}}}

\DeclareMathAlphabet{\mathsfit}{\encodingdefault}{\sfdefault}{m}{sl}
\SetMathAlphabet{\mathsfit}{bold}{\encodingdefault}{\sfdefault}{bx}{n}

\newif\ifsubmit

\submitfalse
\ifsubmit
\newcommand{\bo}[1]{}
\newcommand{\chaowei}[1]{}
\newcommand{\lei}[1]{}
\newcommand{\haonan}[1]{}
\newcommand{\xcyan}[1]{}
\else
\newcommand{\bo}[1]{\textcolor{blue}{Bo: #1}}
\newcommand{\chaowei}[1]{\textcolor{cyan}{Chaowei: #1}}
\newcommand{\lei}[1]{\textcolor{red}{Lei: #1}}
\newcommand{\haonan}[1]{\textcolor{red}{Haonan: #1}}
\newcommand{\xcyan}[1]{\textcolor{red}{TODO: #1}}
\fi

\newcommand{\acAE}{attribute-conditioned adversarial examples\xspace}
\newcommand{\StAdv}{\textit{SemanticAdv}\xspace}
\newcommand{\stAdv}{\textit{semanticAdv}\xspace}

\newcommand{\cutabstractup}{\vspace*{-0.2in}}

\newcommand{\cutsectionup}{\vspace*{-0.12in}}
\newcommand{\cutsectiondown}{\vspace*{-0.12in}}
\newcommand{\cutsubsectionup}{\vspace*{-0.1in}}
\newcommand{\cutsubsectiondown}{\vspace*{-0.07in}}
\newcommand{\cutparagraphup}{\vspace*{-0.12in}}
\newcommand{\cuttablecaptionup}{\vspace*{-0.05in}}
\newcommand{\cuttablecaptiondown}{\vspace*{-0.12in}}
\newcommand{\cutfigurecaptionup}{\vspace*{-0.05in}}
\newcommand{\cutfigurecaptiondown}{\vspace*{-0.15in}}

\usepackage[symbol]{footmisc}

\begin{document}
\pagestyle{headings}
\mainmatter
\def\ECCVSubNumber{2059}  

\title{SemanticAdv: Generating Adversarial Examples \\via Attribute-conditioned Image Editing} 

\titlerunning{SemanticAdv}
\author{
Haonan Qiu\thanks{Alphabetical ordering; The first three authors contributed equally.}\inst{1} \and
Chaowei Xiao$^*$\inst{2} \and
Lei Yang$^*$\inst{3} \and
Xinchen Yan\inst{2,5}\thanks{Work partially done as a PhD student at University of Michigan.} \and \\
Honglak Lee\inst{2} \and
Bo Li\inst{4}
}
\authorrunning{H. Qiu et al.}
\institute{
The Chinese University of Hong Kong, Shenzhen \and
University of Michigan, Ann Arbor \and
The Chinese University of Hong Kong \and
University of Illinois Urbana-Champaign \and
Uber ATG, San Francisco
}

\maketitle

\cutabstractup
\begin{abstract}
Deep neural networks (DNNs) have achieved great successes in various vision applications due to their strong expressive power. 
However, recent studies have shown that DNNs are vulnerable to adversarial examples which are manipulated instances targeting to mislead DNNs to make incorrect predictions. 
Currently, most such adversarial examples try to guarantee ``subtle perturbation" by limiting the $L_p$ norm of the perturbation.
In this paper, we propose \StAdv to generate a new type of \emph{semantically realistic} adversarial examples via attribute-conditioned image editing.
Compared to existing methods, our \StAdv enables fine-grained analysis and evaluation of DNNs with input variations in the attribute space.
We conduct comprehensive experiments to show that our adversarial examples not only exhibit semantically meaningful appearances 
but also achieve high targeted attack success rates under both whitebox and blackbox settings. 
Moreover, we show that the existing pixel-based and attribute-based defense methods fail to defend against  \StAdv. 
We demonstrate the applicability of \emph{SemanticAdv} on both face recognition and general street-view images to show its generalization.
Such non-$L_p$ bounded adversarial examples with controlled attribute manipulation can shed light on further understanding about vulnerabilities of DNNs as well as novel defense approaches. 

\end{abstract}

\cutsectionup
\section{Introduction}
\cutsectiondown

Deep neural networks (DNNs) have demonstrated great successes in advancing the state-of-the-art performance in various vision tasks~\cite{krizhevsky2012imagenet,simonyan2014very,szegedy2015going,he2016deep,schroff2015facenet,long2015fully,Yu2017,chen2017deeplab} and have been widely used in many safety-critical applications such as face verification and autonomous driving~\cite{zhang2018deeproad}.
At the same time, several studies~\cite{szegedy2013intriguing,goodfellow2014explaining,moosavi2016deepfool,papernot2016limitations,carlini2017towards,xiao2018generating,xiao2018spatially,xiao2018characterizing} have revealed the vulnerablity of DNNs against input variations.
For example, carefully crafted $L_p$ bounded perturbations added to the pristine input images can introduce arbitrary prediction errors during testing time.
While being visually imperceptible, $L_p$ bounded adversarial attacks have certain limitations as they only capture the variations in the raw pixel space and cannot guarantee the semantic realism for the generated instances.
Recent works~\cite{xiao2018spatially,kang2019testing,wong2019wasserstein} have shown the limitations of only measuring and evaluating the $L_p$ bounded perturbation (e.g., cannot handle variations in lighting conditions).
Therefore, understanding the failure modes of deep neural networks beyond raw pixel variations including semantic perturbations requires further understanding and exploration.

\begin{figure}[t]
    \centering
    \includegraphics[width=\linewidth]{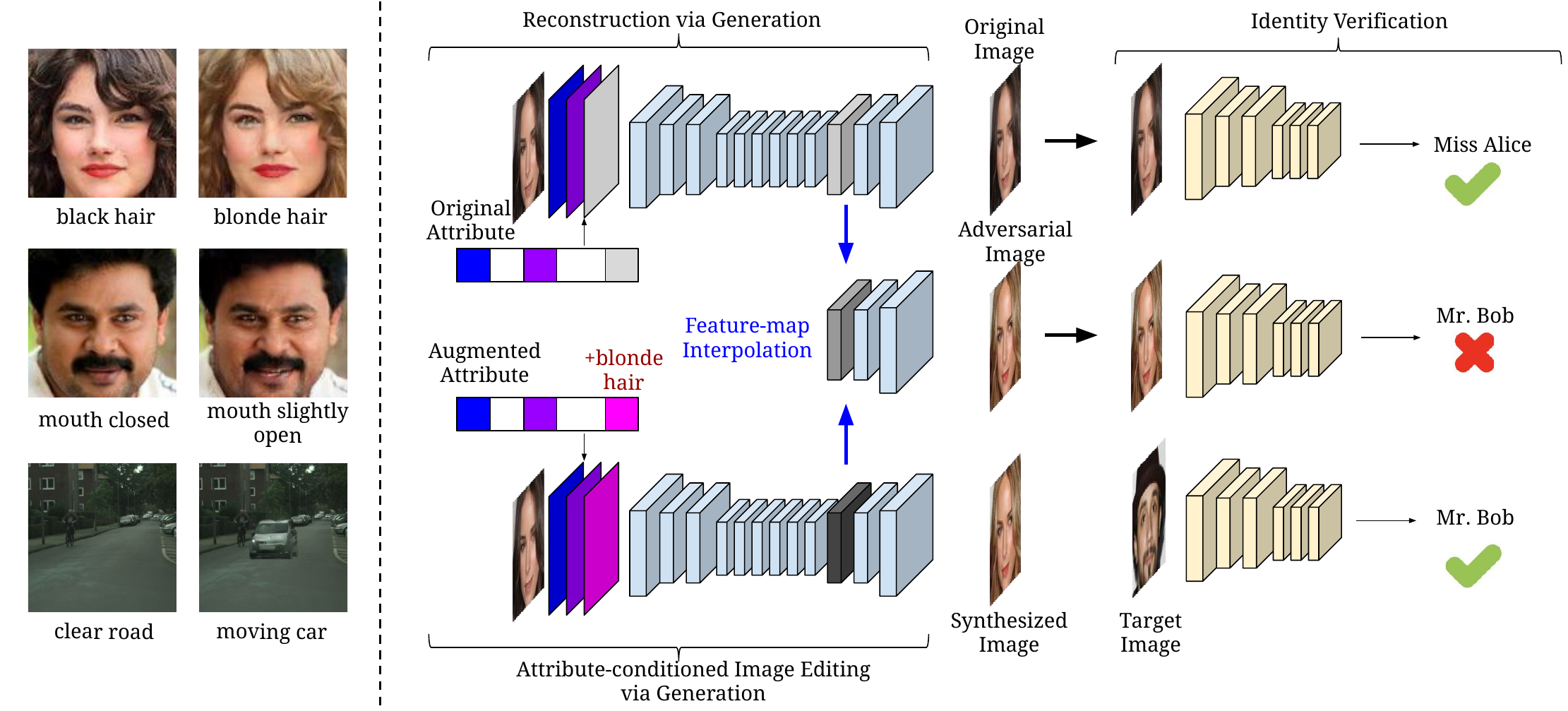}
    \cutfigurecaptionup
    \caption{
    Pipeline of \StAdv Left: Each row shows a pair of images differ in only one semantic aspect.
    One of them is sampled from the ground-truth dataset, while the other one is created by our conditional image generator, which is adversary to the  recognition model (e.g., face identification network and semantic segmentation network).
    Right: Overview of the proposed attribute-conditioned \StAdv against the face identity verification model
    }
    \label{fig:intro}
    \cutfigurecaptiondown
\end{figure}

In this work, we focus on studying how DNNs respond towards semantically meaningful perturbations in the visual attribute space.
In the visual recognition literature, visual attributes~\cite{farhadi2009describing,kumar2009attribute,parikh2011relative} are properties observable in images that have human-designated properties (e.g., \textit{black hair} and \textit{blonde hair}).
As illustrated in Figure~\ref{fig:intro} (left), given an input image with known attributes, we would like to craft semantically meaningful (attribute-conditioned) adversarial examples via image editing along a single attribute or a subset of attributes while keeping the rest unchanged.
Compared to traditional $L_p$ bounded adversarial perturbations or semantic perturbations on global color and texture~\cite{bhattad2020unrestricted}, such attribute-based image editing enables the users to conduct a fine-grained analysis and evaluation of the DNN models through removing one or a set of visual aspects or adding one object into the scene.
We believe our attribute-conditioned image editing is a natural way of introducing semantic perturbations, and it preserves clear interpretability as: wearing a new pair of glasses or having the hair dyed with a different color.

To facilitate the generation of semantic adversarial perturbations along a single attribute dimension, we take advantage of the disentangled representation in deep image generative models~\cite{radford2015unsupervised,karras2017progressive,brock2018large,yan2016attribute2image,choi2018stargan,bau2018gan,yao20183d,johnson2018image}.
Such disentangled representation allows us to explore the variations for a specific semantic factor while keeping the other factors unchanged.
As illustrated in Figure~\ref{fig:intro} (right), we first leverage an attribute-conditioned image editing model~\cite{choi2018stargan} to construct a new instance which is very similar to the source except one semantic aspect (the source image is given as input).
Given such pair of images, we synthesize the adversarial example by interpolating between the pair of images in the \emph{feature-map space}.
As the interpolation is constrained by the image pairs, the appearance of the resulting semantic adversarial example resembles both of them.

To validate the effectiveness of our proposed \StAdv by attribute-conditioned image editing, we consider two real-world tasks, including face verification and landmark detection.
We conduct both qualitative and quantitative evaluations on CelebA dataset~\cite{liu2015deep}. The results show that our \StAdv not only achieves high targeted attack success rate and also preserves the semantic meaning of the corresponding input images.
To further demonstrate the applicability of our \StAdv beyond face domain, we extend the framework to generate adversarial street-view images.
We treat semantic layouts as input attributes and use the layout-conditioned image editing model~\cite{hong2018learning} pre-trained on Cityscape dataset~\cite{cordts2016cityscapes}.
Our results show that a well-trained semantic segmentation model can be successfully attacked to neglect the pedestrian if we insert another object by the side using our image editing model.
In addition, we show that existing adversarial training-based defense method is less effective against our attack method, which motivates further defense strategies against such semantic adversarial examples.

Our contributions are summarized as follows:
%
(1) We propose a novel method \StAdv to generate semantically meaningful adversarial examples via attribute-conditioned image editing based on \textbf{feature-space} interpolation. 
Compared to existing adversarial attacks, our method enables fine-grained attribute analysis as well as further evaluation of vulnerabilities for DNN models. Such semantic adversarial examples also provide explainable analysis for different attributes in terms of their robustness and editing flexibility.
(2) We conduct extensive experiments and show that the proposed feature-space interpolation strategy can generate high quality \acAE more effectively than the simple attribute-space interpolation. 
Additionally, our \StAdv exhibits high attack \textbf{transferability} as well as 67.7\% query-free \textbf{black-box attack} success rate on a real-world face verification platform.
(3) We empirically show that, compared to $L_p$ attacks, the existing per-pixel based as well as attribute-based defense methods fail to defend against our \StAdv, which indicates that such semantic adversarial examples identify certain unexplored vulnerable landscape of DNNs. 
(4) To demonstrate the applicability and generalization of \StAdv beyond the face recognition domain,  we extend the framework to generate adversarial street-view images that fool semantic segmentation models effectively.

\cutsectionup
\section{Related Work} 
\cutsectiondown

\paragraph{Semantic image editing.} 

Semantic image synthesis and manipulation is a popular research topic in machine learning, graphics and vision.
Thanks to recent advances in deep generative models~\cite{kingma2013auto,goodfellow2014generative,oord2016pixel} and the empirical analysis of deep classification networks~\cite{krizhevsky2012imagenet,simonyan2014very,szegedy2015going}, past few years have witnessed tremendous breakthroughs towards high-fidelity pure image generation~\cite{radford2015unsupervised,karras2017progressive,brock2018large}, attribute-to-image generation~\cite{yan2016attribute2image,choi2018stargan}, text-to-image generation~\cite{mansimov2015generating,reed2016generative,van2016conditional,odena2017conditional,zhang2017stackgan,johnson2018image}, and image-to-image translation~\cite{isola2017image,zhu2017unpaired,liu2017unsupervised,wang2018high,hong2018learning}.

\cutparagraphup
\paragraph{Adversarial examples.}
Generating $L_p$ bounded adversarial perturbation has been extensively studied recently~\cite{szegedy2013intriguing,goodfellow2014explaining,moosavi2016deepfool,papernot2016limitations,carlini2017towards,xiao2018generating}. 
To further explore diverse adversarial attacks and potentially help inspire defense mechanisms, it is important to generate the so-called ``unrestricted" adversarial examples which contain unrestricted magnitude of perturbation while still preserve perceptual realism~\cite{brown2018unrestricted}.
Recently, \cite{xiao2018spatially,engstrom2017rotation} propose to spatially transform the image patches instead of adding pixel-wise perturbation, while such spatial transformation does not consider semantic information.
Our proposed \stAdv focuses on generating unrestricted perturbation with semantically meaningful patterns guided by visual attributes.

Relevant to our work, \cite{song2018constructing} proposed to synthesize adversarial examples with an unconditional generative model. 
\cite{bhattad2020unrestricted} studied semantic transformation in only the color or texture space.
Compared to these works, \stAdv is able to generate adversarial examples in a controllable fashion using specific visual attributes by performing manipulation in the feature space.
We further analyze the robustness of the recognition system by generating adversarial examples guided by different visual attributes.
Concurrent to our work, \cite{joshi2019semantic} proposed to generate semantic-based attacks
against a restricted binary classifier, while our attack is able to mislead the model towards arbitrary adversarial targets. 
They conduct the manipulation within the attribution space which is less flexible and effective than our proposed feature-space interpolation.

\cutsectionup
\section{SemanticAdv}
\cutsectiondown

\subsection{Problem Definition}
\cutsubsectiondown

Let $\mathcal{M}$ be a machine learning model trained on a dataset $\mathcal{D} = \left\{ (\mathbf{x},\mathbf{y}) \right\}$ consisting of image-label pairs, where $\mathbf{x} \in \mathbb{R}^{H \times W \times D_I}$ and $\mathbf{y} \in \mathbb{R}^{D_L}$ denote the image and the ground-truth label, respectively.
Here, $H$, $W$, $D_I$, and $D_L$ denote the image height, image width, number of image channels, and label dimensions, respectively.
For each image $\mathbf{x}$, our model $\mathcal{M}$ makes a prediction $\hat{\mathbf{y}} = \mathcal{M}(\mathbf{x}) \in \mathbb{R}^{D_L}$.
%
%
Given a target image-label pair $(\mathbf{x}^\text{tgt},\mathbf{y}^\text{tgt})$ and $\mathbf{y} \neq \mathbf{y}^\text{tgt}$, a \textit{traditional attacker} aims to synthesize adversarial examples $\mathbf{x}^\text{adv}$ by adding pixel-wise perturbations to or spatially transforming the original image $\mathbf{x}$ such that $\mathcal{M}(\mathbf{x}^\text{adv}) = \mathbf{y}^\text{tgt}$.
In this work, we consider a \textit{semantic attacker} that generates semantically meaningful perturbation via attribute-conditioned image editing with a conditional generative model $\mathcal{G}$. 
Compared to the traditional attacker, the proposed attack method generates adversarial examples in a more controllable fashion by editing a single semantic aspect through attribute-conditioned image editing.

\subsection{Attribute-conditioned Image Editing}
In order to produce semantically meaningful perturbations, we first introduce how to synthesize attribute-conditioned images through interpolation.
%


\cutparagraphup

\paragraph{Semantic image editing.}

For simplicity, we start with the formulation where the input attribute is represented as a compact vector.
This formulation can be directly extended to other input attribute formats including semantic layouts.
Let $\mathbf{c} \in \mathbb{R}^{D_C}$ be an attribute representation reflecting the semantic factors (e.g., expression or hair color of a portrait image) of image $\mathbf{x}$, where $D_C$ indicates the attribute dimension and $c_i \in \{0, 1\}$ indicates the existence of $i$-th attribute.
We are interested in performing semantic image editing using the attribute-conditioned image generator $\mathcal{G}$.
%
For example, given a portrait image of a girl with \texttt{black hair} and the new attribute \texttt{blonde hair}, our generator is supposed to synthesize a new image that turns the girl's hair color from black to blonde while keeping the rest of appearance unchanged.
The synthesized image is denoted as $\mathbf{x}^\text{new} = \mathcal{G}(\mathbf{x}, \mathbf{c}^\text{new})$ where $\mathbf{c}^\text{new} \in \mathbb{R}^{D_C}$ is the new attribute.
In the special case when there is no attribute change ($\mathbf{c} = \mathbf{c}^\text{new}$), the generator simply reconstructs the input: $\mathbf{x'} = \mathcal{G}(\mathbf{x}, \mathbf{c})$ (ideally, we hope $\mathbf{x'}$ equals to $\mathbf{x}$).
As our attribute representation is disentangled and the change of attribute value is sufficiently small (e.g., we only edit a single semantic attribute), our synthesized image $\mathbf{x}^\text{new}$ is expected to be close to the data manifold~\cite{bengio2013better,reed2014learning,radford2015unsupervised}.
In addition, we can generate many similar images by linearly interpolating between the image pair $\mathbf{x}$ and $\mathbf{x}^\text{new}$ in the attribute-space or the feature-space of the image-conditioned generator $\mathcal{G}$, which is supported by the previous work~\cite{yan2016attribute2image,radford2015unsupervised,bau2018gan}

\cutparagraphup
\paragraph{Attribute-space interpolation.} 
%
Given a pair of attributes $\mathbf{c}$ and $\mathbf{c}^\text{new}$, we introduce an interpolation parameter $\alpha \in (0,1)$ to generate the augmented attribute vector $\mathbf{c}^{*} \in \mathbb{R}^{D_C}$ (see Eq.~\ref{eqn:method-attr-interp}).
Given augmented attribute $\mathbf{c}^*$ and original image $\mathbf{x}$, we produce the image $\mathbf{x}^\text{*}$ by the generator $\mathcal{G}$ through attribute-space interpolation.
%
%
\begin{align}
\mathbf{x}^\text{*} &= \mathcal{G}(\mathbf{x}, \mathbf{c}^\text{*})\nonumber \\
\mathbf{c}^\text{*} &= \alpha \cdot \mathbf{c} + (1 - \alpha) \cdot \mathbf{c}^\text{new} \text{, where $\alpha \in [0, 1]$}
\label{eqn:method-attr-interp}
\end{align}

\cutparagraphup
\paragraph{Feature-map interpolation.} Alternatively, we propose to interpolate using the feature map produced by the generator $\mathcal{G} = \mathcal{G}_\text{dec} \circ \mathcal{G}_\text{enc}$. 
Here, $\mathcal{G}_\text{enc}$ is the encoder module that takes the image as input and outputs the feature map.
Similarly, $\mathcal{G}_\text{dec}$ is the decoder module that takes the feature map as input and outputs the synthesized image.
Let $\mathbf{f^*} = \mathcal{G}_\text{enc}(\mathbf{x}, \mathbf{c}) \in \mathbb{R}^{H_F \times W_F \times C_F}$ be the feature map of an intermediate layer in the generator, where $H_F$, $W_F$ and $C_F$ indicate the height, width, and number of channels in the feature map.
\begin{align}
\mathbf{x}^\text{*} &= \mathcal{G}_\text{dec}(\mathbf{f}^\text{*})\nonumber\\
\mathbf{f}^\text{*} &= \pmb{\beta} \odot \mathcal{G}_\text{enc}(\mathbf{x}, \mathbf{c})
+ (\mathbf{1} - \pmb{\beta}) \odot \mathcal{G}_\text{enc}(\mathbf{x}, \mathbf{c}^\text{new})
\label{eqn:method-feature-interp}
\end{align}

Compared to the attribute-space interpolation which is parameterized by a scalar $\alpha$, we parameterize feature-map interpolation by a tensor ${\pmb{\beta}} \in \mathbb{R}^{H_F \times W_F \times C_F}$ ($\beta_{h,w,k} \in [0, 1]$, where $1\leq h \leq H_F$, $1 \leq w \leq W_F$, and $1 \leq k \leq C_F$) with the same shape as the feature map.
Compared to linear interpolation over attribute-space, such design introduces more flexibility for adversarial attacks.
Empirical results in Section~\ref{sec:compare} show such design is critical to maintain both attack success and good perceptual quality at the same time.

\cutsubsectionup
\subsection{Generating Semantically Meaningful Adversarial Examples}
\cutsubsectiondown
Existing work obtains the adversarial image $\mathbf{x}^\text{adv}$ by adding perturbations or transforming the input image $\mathbf{x}$ directly.
In contrast, our semantic attack method requires additional attribute-conditioned image generator $\mathcal{G}$ during the adversarial image generation through interpolation.
As we see in Eq.~\ref{eqn:method-full-objective}, the first term of our objective function is the adversarial metric, the second term is a smoothness constraint to guarantee the perceptual quality, and $\lambda$ is used to control the balance between the two terms.
The adversarial metric is minimized once the model $\mathcal{M}$ has been successfully attacked towards the target image-label pair $(\mathbf{x}^\text{tgt}, \mathbf{y}^\text{tgt})$.
For identify verification, $\mathbf{y}^\text{tgt}$ is the identity representation of the target image; For structured prediction tasks in our paper, $\mathbf{y}^\text{tgt}$ either represents certain coordinates (landmark detection) or semantic label maps (semantic segmentation).

\begin{align}
\mathbf{x}^\text{adv} 
&= {\argmin}_{\mathbf{x}^*} \mathcal{L}(\mathbf{x}^*) \nonumber\\
\mathcal{L}(\mathbf{x}^*) &= \mathcal{L}_\text{adv}(\mathbf{x}^*; \mathcal{M}, \mathbf{y}^\text{tgt}) + \lambda \cdot \mathcal{L}_\text{smooth}(\mathbf{x}^*)
\label{eqn:method-full-objective}
\end{align}

\cutparagraphup
\paragraph{Identity verification.}
In the identity verification task, two images are considered to be the same identity if the corresponding identity embeddings from the verification model $\mathcal{M}$ are reasonably close.
\begin{equation}
\mathcal{L}_\text{adv}(\mathbf{x}^* ; \mathcal{M}, \mathbf{y}^\text{tgt}) = \max\{\kappa, \Phi_\mathcal{M}^\text{id}(\mathbf{x}^*, \mathbf{x}^\text{tgt})\}
\label{eqn:method-id-objective}
\end{equation}

As we see in Eq.~\ref{eqn:method-id-objective}, $\Phi_\mathcal{M}^\text{id}(\cdot, \cdot)$ measures the distance between two identity embeddings from the model $\mathcal{M}$, where the normalized $L_2$ distance is used in our setting. 
In addition, we introduce the parameter $\kappa$ representing the constant related to the false positive rate (FPR) threshold computed from the development set.

\cutparagraphup
\paragraph{Structured prediction.}
For structured prediction tasks such as landmark detection and semantic segmentation, we use
Houdini objective proposed in~\cite{cisse2017houdini} as our adversarial metric and select the target landmark (semantic segmentation) target as  $\mathbf{y}^\text{tgt}$. 
As we see in the equation, $\Phi_\mathcal{M}(\cdot, \cdot)$ is a scoring function for each image-label pair and $\gamma$ is the threshold.
In addition,
$l(\mathbf{y}^*,\mathbf{y}^\text{tgt})$ is task loss decided by the specific adversarial target, where $\mathbf{y}^* = \mathcal{M}(\mathbf{x}^*)$.

\begin{align}
\mathcal{L}_\text{adv}(\mathbf{x}^*
; \mathcal{M}, \mathbf{y}^\text{tgt}) =& P_{\gamma \sim \mathcal{N}(0,1)}
\Big[ \Phi_\mathcal{M}(\mathbf{x}^*
, \mathbf{y}^*) - \Phi_\mathcal{M}(\mathbf{x}^*
, \mathbf{y}^\text{tgt})<\gamma\Big] \cdot l(\mathbf{y}^*,\mathbf{y}^\text{tgt})
\end{align}

\cutparagraphup
\paragraph{Interpolation smoothness $\mathcal{L}_\text{smooth}$.}
As the tensor to be interpolated in the feature-map space has far more parameters compared to the attribute itself, we propose to enforce a smoothness constraint on the tensor $\alpha$ used in feature-map interpolation.
As we see in Eq.~\ref{eqn:method-smoothness}, the smoothness loss encourages the interpolation tensors to consist of piece-wise constant patches spatially, which has been widely used as a pixel-wise de-noising objective for natural image processing~\cite{mahendran2015understanding,johnson2016perceptual}.
\begin{equation}
\mathcal{L}_\text{smooth}(\pmb{\beta}) = \sum_{h=1}^{H_F-1} \sum_{w=1}^{W_F} \| {\pmb{\beta}}_{h+1,w} - {\pmb{\beta}}_{h,w}\|^2_2
+ \sum_{h=1}^{H_F} \sum_{w=1}^{W_F-1} \|{\pmb{\beta}}_{h,w+1} - {\pmb{\beta}}_{h,w}\|^2_2
\label{eqn:method-smoothness}
\end{equation}


\cutsectionup
\section{Experiments}
\cutsectiondown

In the experimental section, we mainly focus on analyzing the proposed \StAdv in attacking state-of-the-art face recognition systems~\cite{sun2014deep,schroff2015facenet,zhang2018accelerated,wang2018cosface} due to its wide applicability (e.g., identification for mobile payment) in the real world.
We attack both face verification and face landmark detection by generating attribute-conditioned adversarial examples using annotations from CelebA dataset~\cite{liu2015deep}.
In addition, we extend our attack to urban street scenes with semantic label maps as the condition.
We attack the semantic segmentation model DRN-D-22~\cite{Yu2017} previously trained on Cityscape~\cite{cordts2016cityscapes} by generating adversarial examples with dynamic objects manipulated (e.g., insert a car into the scene).

The experimental section is organized as follows. 
First, we analyze the quality of generated adversarial examples and qualitatively compare our method with $\mathcal{L}_p$ bounded pixel-wise optimization-based methods~\cite{carlini2017towards,dong2018boosting,xie2019improving}.
Second, we provide both qualitative and quantitative results by controlling single semantic attribute.
In terms of attack transferability, we evaluate our proposed \StAdv in various settings and further demonstrate the effectiveness of our method via query-free black-box attacks against online face verification platforms.
Third, we compare our method with the baseline methods against different defense methods on the face verification task.
Fourth, we demonstrate that our \StAdv is a general framework by showing the results in other tasks including face landmark detection and street-view semantic segmentation.

\cutsubsectionup
\subsection{Experimental Setup}\label{sec:expsetup}

\cutparagraphup
\paragraph{Face identity verification.} 
We select \texttt{ResNet-50} and \texttt{ResNet-101}~\cite{he2016deep} trained on MS-Celeb-1M~\cite{guo2016ms,deng2019arcface} as our face verification models.
The models are trained using two different objectives, namely, \texttt{softmax} loss~\cite{sun2014deep,zhang2018accelerated} and \texttt{cosine} loss~\cite{wang2018cosface}.
For simplicity, we use the notation ``R-N-S'' to indicate the model with $N$-layer residual blocks as backbone trained using \texttt{softmax} loss, while ``R-N-C'' indicates the same backbone trained using \texttt{cosine} loss.
The distance between face features is measured by normalized L2 distance.
For R-101-S model, we decide the parameter $\kappa$ based on the false positive rate (FPR) for the identity verification task.
Four different FPRs have been used: $10^{-3}$ (with $\kappa = 1.24$), $3 \times 10^{-4}$ (with $\kappa = 1.05$), $10^{-4}$ (with $\kappa = 0.60$), and $<10^{-4}$ (with $\kappa = 0.30$).
The distance metrics and selected thresholds are commonly used when evaluating the performance of face recognition model~\cite{klare2015pushing,kemelmacher2016megaface}.
Supplementary provides more details on the performance of face recognition models and their corresponding $\kappa$.
To distinguish between the FPR we used in generating adversarial examples and the other FPR used in evaluation, we introduce two notations ``Generation FPR (G-FPR)'' and ``Test FPR (T-FPR)''.
For the experiment with query-free black-box API attacks, we use two online face verification services provided by Face++~\cite{faceplusplus} and AliYun~\cite{aliyun}.

\cutparagraphup
\paragraph{Semantic attacks on face images.}
In our experiments, we randomly sample $1,280$ distinct identities form CelebA~\cite{liu2015deep} and use the StarGAN~\cite{choi2018stargan} for attribute-conditional image editing.
In particular, we re-train our model on CelebA by aligning the face landmarks and then resizing images to resolution $112 \times 112$. 
We select 17 identity-preserving attributes as our analysis, as such attributes mainly reflect variations in facial expression and hair color.

In feature-map interpolation, to reduce the reconstruction error brought by the generator (e.g., $\mathbf{x} \neq \mathcal{G}(\mathbf{x}, \mathbf{c})$) in practice, we take one more step to obtain the updated feature map $\mathbf{f}^\prime = \mathcal{G}_\text{enc}(\mathbf{x}^\prime, \mathbf{c})$, where $\mathbf{x}^\prime = \argmin_{\mathbf{x}^\prime} \| \mathcal{G}(\mathbf{x}^\prime, \mathbf{c}) - \mathbf{x}\|$.
%
%

For each distinct identity pair $(\mathbf{x}, \mathbf{x}^\text{tgt})$,
we perform \stAdv guided by each of the 17 attributes (e.g., we intentionally add or remove one specific attribute while keeping the rest unchanged). 
In total, for each image $\mathbf{x}$, we generate 17 adversarial images with different augmented attributes.
In the experiments, we select a commonly-used pixel-wise adversarial attack method~\cite{carlini2017towards} (referred as CW) as our baseline. 
Compared to our proposed method, CW does not require visual attributes as part of the system, as it only generates one adversarial example for each instance.
We refer the corresponding attack success rate as the instance-wise success rate in which the attack success rate is calculated for each instance.
For each instance with 17 adversarial images using different augmented attributes, if one of the 17 produced images can attack successfully, we count the attack of this instance as one success, vice verse.

\cutparagraphup
\paragraph{Face landmark detection.}
We select Face Alignment Network (FAN)~\cite{bulat2017far} trained on 300W-LP~\cite{zhu2016face} and fine-tuned on 300-W~\cite{sagonas2013300} for 2D landmark detection. 
The network is constructed by stacking Hour-Glass networks~\cite{newell2016stacked} with hierarchical block~\cite{bulat2017binarized}.
Given a face image as input, FAN outputs 2D heatmaps which can be subsequently leveraged to yield $68$ 2D landmarks.

\cutparagraphup
\paragraph{Semantic attacks on street-view images.}
We select DRN-D-22~\cite{Yu2017} as our semantic segmentation model and fine-tune the model on image regions with resolution $256 \times 256$.
To synthesize semantic adversarial perturbations, we consider semantic label maps as the input attribute and leverage a generative image manipulation model~\cite{hong2018learning} pre-trained on CityScape~\cite{cordts2016cityscapes} dataset.
Given an input semantic label map at resolution $256 \times 256$, we select a target object instance (e.g., a pedestrian) to attack.
Then, we create a manipulated semantic label map by inserting another object instance (e.g., a car) in the vicinity of the target object.
Similar to the experiments in the face domain, for both semantic label maps, we use the image manipulation encoder to extract features (with $1,024$ channels at spatial resolution $16\times16$) and conduct feature-space interpolation.
We synthesize the final image by feeding the interpolated features to the image manipulation decoder.
By searching the interpolation coefficient that maximizes the attack rate, we are able to fool the segmentation model with the synthesized final image.

\cutsubsectionup
\subsection{\StAdv on Face Identity Verification}
\label{sec:compare}

\begin{figure}[t]
	\centering
	\includegraphics[width=0.9\linewidth]{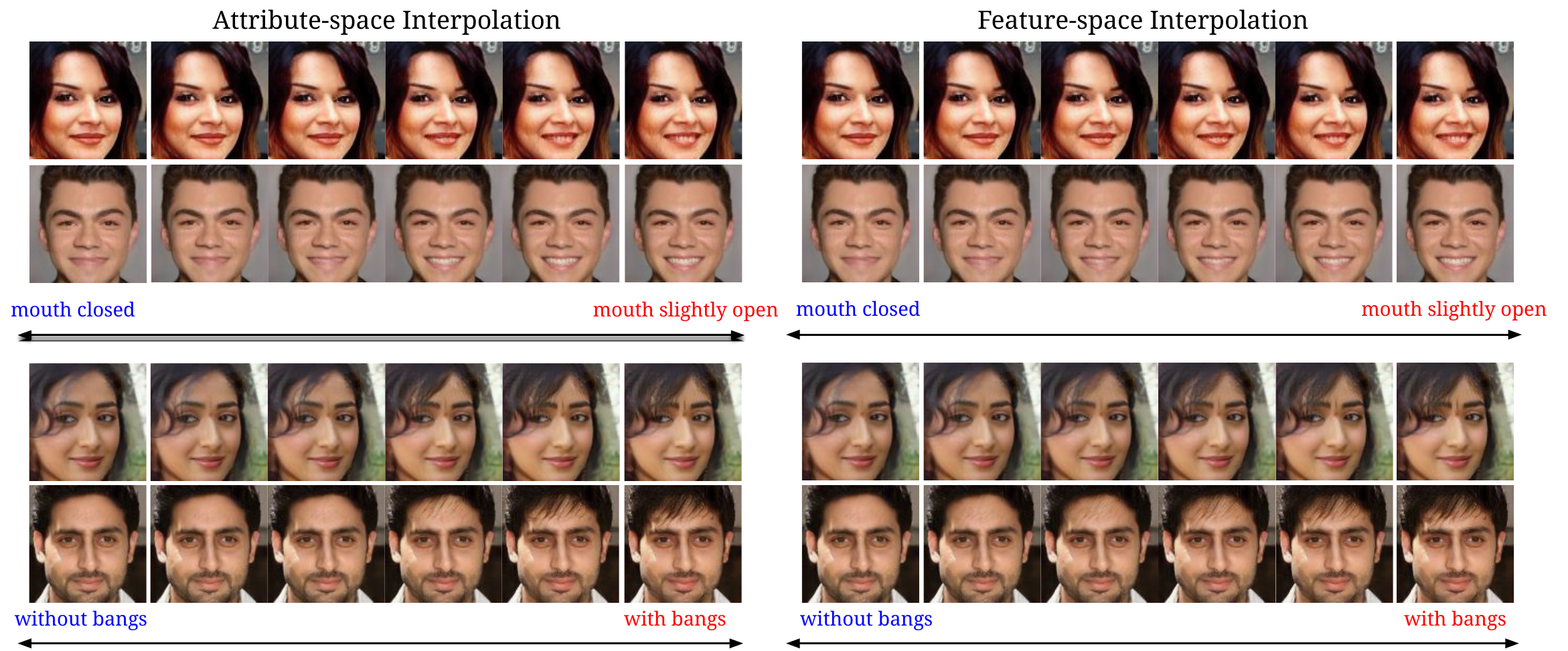}
	\cutfigurecaptionup
	\caption{Qualitative comparisons between attribute-space and feature-space interpolation
	In our visualization, we set the interpolation parameter to be $0.0, 0.2, 0.4, 0.6, 0.8, 1.0$
	}
	\label{fig:figure_appendix_interp}
\end{figure}

\begin{table}[ht]
    \centering
    \caption{Attack success rate by selecting attribute or different layer's feature-map for interpolation on R-101-S(\%) using $\text{G-FPR} = \text{T-FPR} = 10^{-3}$.
    Here, $\mathbf{f}_{i}$ indicates the feature-map after $i$-th \texttt{up-sampling} operation.
    $\mathbf{f}_{-2}$ and $\mathbf{f}_{-1}$ are the first and the second feature-maps after the last \texttt{down-sampling} operation, respectively.}
    \scalebox{0.95}{
        \begin{tabular}{l|c|c|c|c|c|c}
        \hline
       \multicolumn{1}{c|}{\multirow{2}{*}{Interpolation / Attack Success (\%)}} & \multicolumn{5}{c|}{Feature} &  \multicolumn{1}{c}{\multirow{2}{*}{Attribute} }\\
       \cline{2-6}
        & $\mathbf{f}_{-2} $ & $\mathbf{f}_{-1}$ & $\mathbf{f}_0$ & $\mathbf{f}_1$ & $\mathbf{f}_2$ \\
        \hline\hline
       $\mathbf{x}^\text{adv}$, G-FPR = $10^{-3}$ & 99.38 & 100.00 & \textbf{100.00} & 100.00 & 99.69 & 0.08 \\
       $\mathbf{x}^\text{adv}$, G-FPR = $10^{-4}$ & 59.53 & 98.44 & \textbf{99.45} & 97.58 & 73.52 & 0.00 \\
        \hline
        \end{tabular}
    }
    \label{tab:attack_layer}
    \cutfigurecaptiondown
\end{table}

\cutparagraphup
\paragraph{Attribute-space vs. feature-space interpolation.}
First, we qualitatively compare the two interpolation methods and found that both attribute-space and feature-space interpolation can generate reasonably looking samples (see Figure~\ref{fig:figure_appendix_interp}) through interpolation (these are not adversarial examples).
However, we found the two interpolation methods perform differently when we optimize using the adversarial objective (Eq.~\ref{eqn:method-full-objective}).
We measure the attack success rate of attribute-space interpolation (with G-FPR = T-FPR = $10^{-3}$): $0.08\%$ on R-101-S, $0.31\%$ on R-101-C, and $0.16\%$ on both R-50-S and R-50-C, which consistently fails to attack the face verification model.
%
Compared to attribute-space interpolation, generating adversarial examples with feature-space interpolation produces much better quantitative results (see Table~\ref{tab:attack_layer}).
We conjecture that this is because the high dimensional feature space can provide more manipulation freedom.
This also explains one potential reason of poor samples (e.g., blurry with many noticeable artifacts) generated by the method proposed in \cite{joshi2019semantic}.
We select $\mathbf{f}_0$, the last \texttt{conv} layer before \texttt{up-sampling} layer in the generator for feature-space interpolation due to its good performance.

\begin{figure}[th]
    \centering
    \includegraphics[width=0.85\linewidth]{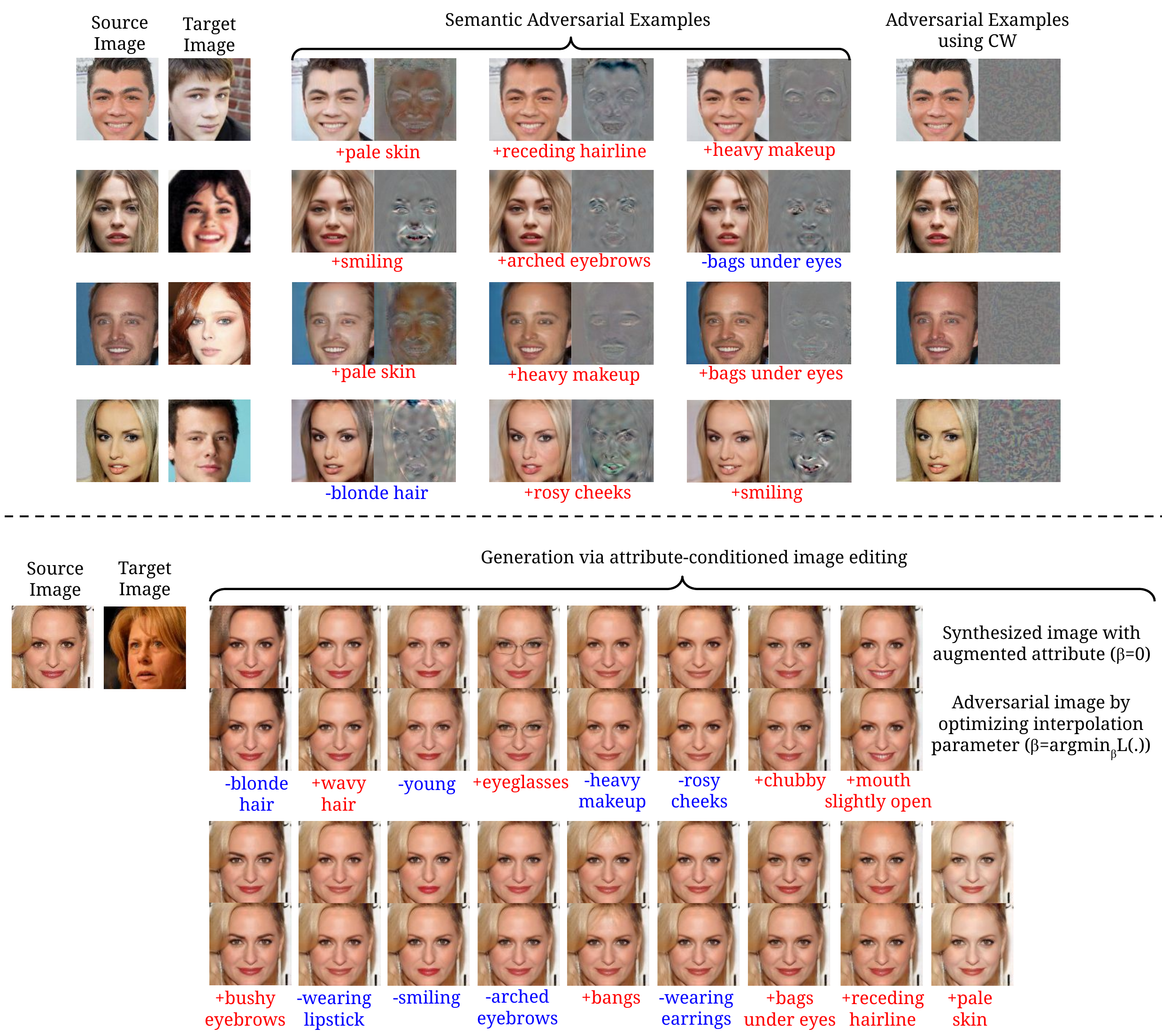}
    \cutfigurecaptionup
    \caption{
    Top: Qualitative comparisons between our proposed \StAdv and pixel-wise adversarial examples generated by CW~\cite{carlini2017towards}. Along with the adversarial examples, we also provide the corresponding perturbations (residual) on the right.
    Perturbations generated by our \StAdv (G-FPR = $10^{-3}$) are unrestricted with semantically meaningful patterns. 
    Bottom: Qualitative analysis on single-attribute adversarial attack (G-FPR = $10^{-3}$).
    More results are shown in the supplementary
    }
    \label{fig:overal_attack_compare}
    \cutfigurecaptiondown
\end{figure}

\paragraph{Qualitative analysis.} 
Figure~\ref{fig:overal_attack_compare} (top) shows the generated adversarial images and corresponding perturbations against R-101-S of \StAdv and CW respectively. The text below each figure is the name of an augmented attribute, the sign before the name represents ``adding'' (in red) or ``removing'' (in blue) the corresponding attribute from the original image.
Figure~\ref{fig:overal_attack_compare} (bottom) shows the adversarial examples with 17 augmented semantic attributes, respectively. 
The attribute names are shown in the bottom. The first row contains images generated by $\mathcal{G}(\mathbf{x}, \mathbf{c}^{\text{new}})$ with an augmented attribute $\mathbf{c}^{\text{new}}$ and the second row includes the corresponding adversarial images under feature-space interpolation. 
It shows that our \StAdv can generate examples with reasonably-looking appearance guided by the corresponding attribute.
In particular, \StAdv is able to generate perturbations on the corresponding regions correlated with the augmented attribute, while the perturbations of CW have no specific pattern and are evenly distributed across the image.

To further measure the perceptual quality of the adversarial images generated by \StAdv in the most strict settings (G-FPR $ < 10^{-4}$), we conduct a user study using Amazon Mechanical Turk (AMT). 
%
%
In total, we collect $2,620$ annotations from $77$ participants.
In $39.14\pm1.96 \%$ (close to random guess $50\%$) of trials, the adversarial images generated by our \StAdv are selected as reasonably-looking images, while $30.27\pm1.96\%$ of trials by CW are selected as reasonably-looking. 
It indicates that \StAdv can generate more perceptually plausible adversarial examples compared with CW under the most strict setting (G-FPR $ < 10^{-4}$).
The corresponding images are shown in supplementary materials.
%

\begin{figure}[h]
    \centering
    \includegraphics[width=0.9\linewidth]{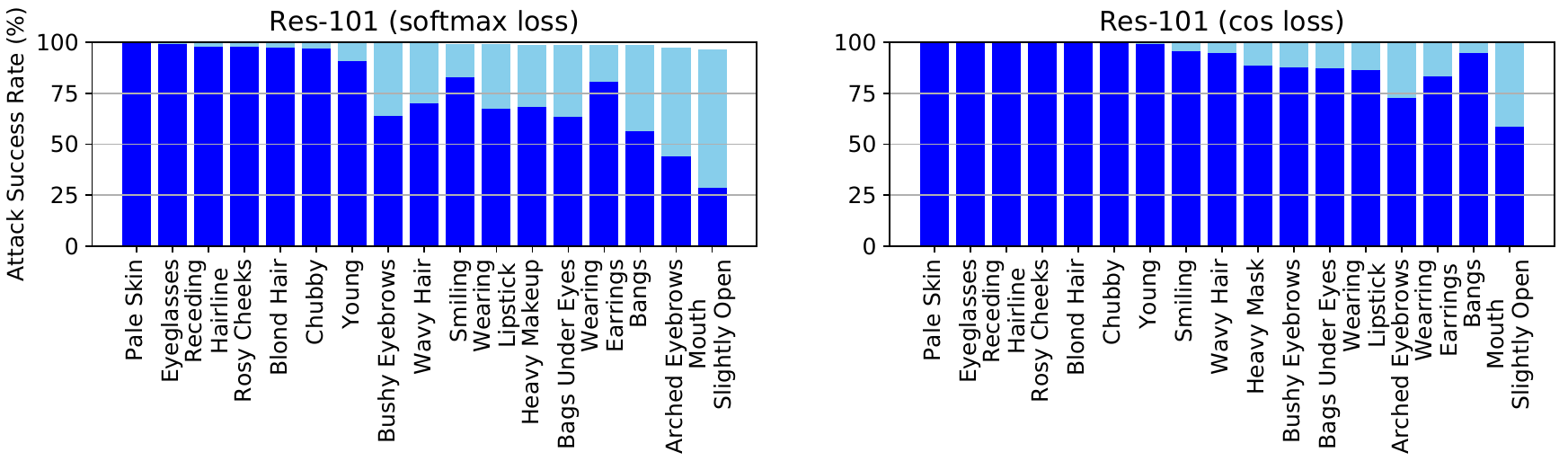}
    \cutfigurecaptionup
    \caption{
    Quantitative analysis on the attack success rate with different single-attribute attacks. In each figure, we show the results correspond to a larger FPR (G-FPR = T-FPR = $10^{-3}$) in \texttt{skyblue} and the results correspond to a smaller FPR (G-FPR = T-FPR = $10^{-4}$) in \texttt{blue}, respectively
    }
    \label{fig:single_attribute_attack_bar}
    \cutfigurecaptiondown
\end{figure}


\cutparagraphup
\paragraph{Single attribute analysis.} 

One of the key advantages of our \StAdv is that we can generate adversarial perturbations in a more controllable fashion guided by the selected semantic attribute.
This allows analyzing the robustness of a recognition system against different types of semantic attacks.
We group the adversarial examples by augmented attributes in various settings.
In Figure~\ref{fig:single_attribute_attack_bar}, we present the attack success rate against two face verification models, namely, R-101-S and R-101-C, using different attributes.
We highlight the bar with light blue for G-FPR = $10^{-3}$ and blue for G-FPR = $10^{-4}$, respectively.
As shown in Figure~\ref{fig:single_attribute_attack_bar}, with a larger T-FPR = $10^{-3}$, our \StAdv can achieve almost 100\% attack success rate across different attributes. 
With a smaller T-FPR = $10^{-4}$, we observe that \StAdv guided by some attributes such as \texttt{Mouth Slightly Open} and \texttt{Arched Eyebrows} achieve less than 50\% attack success rate, while other attributes such as \texttt{Pale Skin} and \texttt{Eyeglasses} are relatively less affected. 
In summary, the above experiments indicate that \StAdv guided by attributes describing the local shape (e.g., mouth, earrings) achieve a relatively lower attack success rate compared to attributes relevant to the color (e.g., hair color) or entire face region (e.g., skin).
This suggests that the face verification models used in our experiments are more robustly trained in terms of detecting local shapes compared to colors.
In practice, we have the flexibility to select attributes for attacking an image based on the perceptual quality and attack success rate.

\cutparagraphup
\paragraph{Transferability analysis.}

To generate adversarial examples under black-box setting, we analyze the transferability of \StAdv in various settings. 
For each model with different FPRs, we select the successfully attacked adversarial examples from Section~\ref{sec:expsetup} to construct our evaluation dataset and evaluate these adversarial samples across different models. 
Table~\ref{tab:transfer_analysis_appendix}(a) illustrates the transferability of \StAdv among different models by using the same FPRs (G-FPR = T-FPR = $10^{-3}$).
Table~\ref{tab:transfer_analysis_appendix}(b) illustrates the result with different FPRs for generation and evaluation (G-FPR = $10^{-4}$ and T-FPR = $10^{-3}$). 
As shown in Table~\ref{tab:transfer_analysis_appendix}(a), adversarial examples generated against models trained with \texttt{softmax} loss exhibit certain transferability compared to models trained with \texttt{cosine} loss. We conduct the same experiment by generating adversarial examples with CW and found it has weaker transferability compared to our \StAdv (results in brackets of Table~\ref{tab:transfer_analysis_appendix}).

As Table~\ref{tab:transfer_analysis_appendix}(b) illustrates, the adversarial examples generated against the model with smaller G-FPR $= 10^{-4}$ exhibit strong attack success rate when evaluating the model with larger T-FPR $=10^{-3}$. 
Especially, we found the adversarial examples generated against R-101-S have the best attack performance on other models.
These findings motivate the analysis of the query-free black-box API attack detailed in the following paragraph.

\begin{table}[t]
    \centering
    \caption{Transferability of \StAdv: cell $(i,j)$ shows attack success rate of adversarial examples generated against $j$-th model and evaluate on $i$-th model. Results of CW are listed in brackets. Left: Results generated with G-FPR = $10^{-3}$ and T-FPR = $10^{-3}$;  Right: Results generated with G-FPR = ${10}^{-4}$ and T-FPR = ${10}^{-3}$}
    \begin{minipage}{0.58\textwidth}
    \centering
    \scalebox{0.65}{\begin{tabular}{l|c|c|c|c}
        \hline
        $\mathcal{M}_\text{test}$ $/$ $\mathcal{M}_\text{opt}$ & R-50-S & R-101-S & R-50-C & R-101-C \\
        \hline\hline
        R-50-S & 1.000 (1.000)& \textbf{0.108} (0.032)& \textbf{0.023} (0.007)& \textbf{0.018} (0.005)\\
        R-101-S & \textbf{0.169} (0.029)& 1.000 (1.000)& \textbf{0.030} (0.009)& \textbf{0.032} (0.011)\\
        R-50-C & \textbf{0.166} (0.054)& \textbf{0.202} (0.079)& 1.000 (1.000)& \textbf{0.048} (0.020)\\
        R-101-C & \textbf{0.120} (0.034)& \textbf{0.236} (0.080)& \textbf{0.040} (0.017)& 1.000 (1.000)\\
        \hline
    \end{tabular}} \\
    (a)
    \end{minipage}
    \begin{minipage}{0.38\textwidth}
    \centering
    \scalebox{0.65}{\begin{tabular}{l|c|c}
        \hline
        $\mathcal{M}_\text{test}$ $/$ $\mathcal{M}_\text{opt}$ & R-50-S & R-101-S \\
        \hline\hline
        R-50-S & 1.000 (1.000) & \textbf{0.862} (0.530) \\
        R-101-S & \textbf{0.874} (0.422) & 1.000 (1.000) \\
        R-50-C & \textbf{0.693} (0.347) & \textbf{0.837} (0.579) \\
        R-101-C & \textbf{0.617} (0.218) & \textbf{0.888} (0.617) \\
        \hline
    \end{tabular}} \\
    (b)
    \end{minipage}
    \label{tab:transfer_analysis_appendix}
    \cuttablecaptiondown
\end{table}

\cutparagraphup
\paragraph{Query-free black-box API attack.}
In this experiment, we generate adversarial examples against R-101-S with G-FPR $ = 10^{-3} (\kappa = 1.24)$, G-FPR $ = 10^{-4} (\kappa=0.60)$, and G-FPR $ < 10^{-4} (\kappa = 0.30) $, respectively.
We evaluate our algorithm on two industry level face verification APIs, namely, Face++ and AliYun.
Since attack transferability has never been explored in concurrent work that generates semantic adversarial examples, we use $\mathcal{L}_p$ bounded pixel-wise methods (CW~\cite{carlini2017towards}, MI-FGSM\cite{dong2018boosting}, M-DI$^2$-FGSM\cite{xie2019improving}) as our baselines. 
We also introduce a much strong baseline by first performing attribute-conditioned image editing and running CW attack on the editted images, which we refer as and StarGAN+CW.
Compared to CW, the latter two devise certain techniques to improve their transferability.
We adopt the ensemble version of MI-FGSM\cite{dong2018boosting} following the original paper.
As shown in Table~\ref{tab:blackbox_analysis}, our proposed \StAdv achieves a much higher attack success rate than the baselines in both APIs under all FPR thresholds (e.g., our adversarial examples generated with G-FPR $< 10^{-4}$ achieves $67.69\%$ attack success rate on Face++ platform with T-FPR $=10^{-3}$).
In addition, we found that lower G-FPR can achieve higher attack success rate on both APIs within the same T-FPR (see our supplementary material for more details). 

\begin{table}[ht]
    \caption{Quantitative analysis on query-free black-box attack. 
    We use ResNet-101 optimized with \texttt{softmax} loss for evaluation and report the attack success rate(\%) on two online face verification platforms. Note that for PGD-based attacks, we adopt MI-FGSM ($\epsilon = 8$) in \cite{dong2018boosting} and M-DI$^2$-FGSM ($\epsilon = 8$) in \cite{xie2019improving}, respectively. For CW, StarGAN+CW and \StAdv, we generate adversarial samples with G-FPR $< 10^{-4}$}
    \centering\small
    \begin{tabular}{l|c|c|c|c}
        \hline
        API name & \multicolumn{2}{c|}{Face++} & \multicolumn{2}{c}{AliYun} \\ \cline{2-5}
        Attacker $/$ Metric & T-FPR = $10^{-3}$ & T-FPR = $10^{-4}$ & T-FPR = $10^{-3}$ & T-FPR = $10^{-4}$\\
        \hline\hline
        CW~\cite{carlini2017towards} & 37.24 & 20.41 & 18.00 & 9.50 \\
        StarGAN+CW & 47.45 & 26.02 & 20.00 & 8.50 \\
        \hline
        MI-FGSM~\cite{dong2018boosting} & 53.89 & 30.57 & 29.50 & 17.50 \\
        M-DI$^2$-FGSM~\cite{xie2019improving} & 56.12 & 33.67 & 30.00 & 18.00 \\
        \hline
        \StAdv & \textbf{67.69} & \textbf{48.21} & \textbf{36.50} & \textbf{19.50} \\
        \hline
    \end{tabular}
    \label{tab:blackbox_analysis}
    \cuttablecaptiondown
\end{table}

\vspace*{-0.05in}

\begin{figure*}[ht]
    \centering
    \includegraphics[width=0.9\linewidth]{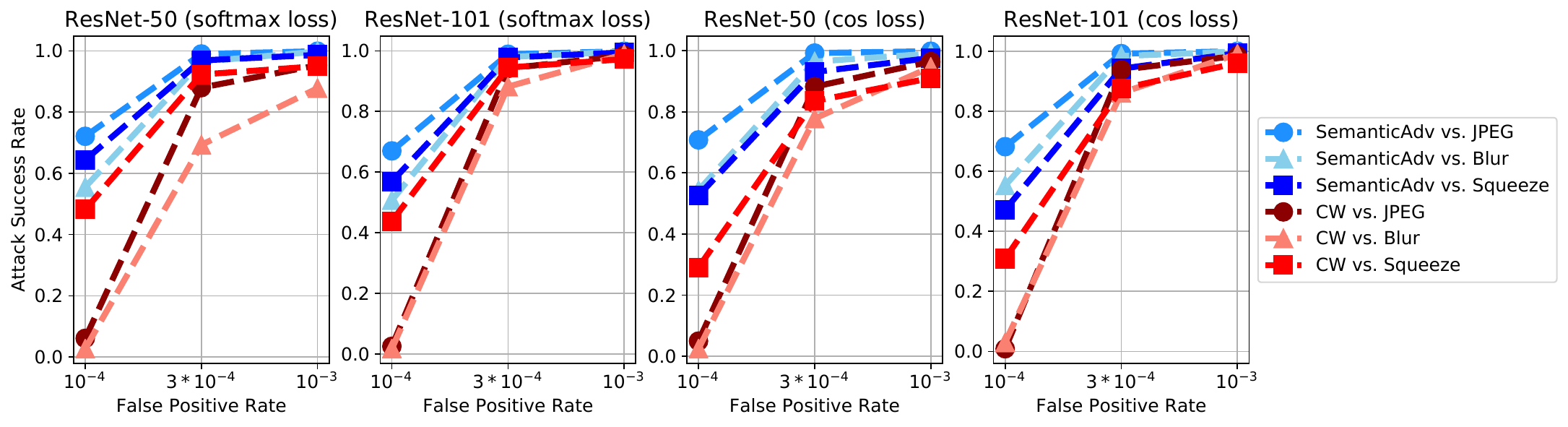}
    \cutfigurecaptionup
    \caption{Quantitative analysis on attacking several defense methods including \texttt{JPEG}~\cite{dziugaite2016study}, \texttt{Blurring}~\cite{li2017adversarial}, and \texttt{Feature Squeezing}~\cite{xu2017feature}}
    \label{fig:defense_tpr_fpr}
    \cutfigurecaptiondown
\end{figure*}

\cutparagraphup
\paragraph{\StAdv against defense methods.}
We evaluate the strength of the proposed attack by testing against five existing defense methods, namely,
\texttt{Feature squeezing}~\cite{xu2017feature}, \texttt{Blurring}~\cite{li2017adversarial}, \texttt{JPEG}~\cite{dziugaite2016study}, \texttt{AMI}~\cite{tao2018attacks} and adversarial training~\cite{madry2017towards}.

Figure~\ref{fig:defense_tpr_fpr} illustrates \StAdv is more robust against the pixel-wise defense methods comparing with CW.
The same G-FPR and T-FPR are used for evaluation. 
Both \StAdv and CW achieve a high attack success rate when T-FPR = $10^{-3}$, while \StAdv marginally outperforms CW when T-FPR goes down to $10^{-4}$.
While defense methods have proven to be effective against CW attacks on classifiers trained with ImageNet~\cite{krizhevsky2012imagenet}, our results indicate that these methods are still vulnerable in the face verification system with small G-FPR.

We further evaluate \StAdv on attribute-based defense method \texttt{AMI}~\cite{tao2018attacks} by constructing adversarial examples for the pretrained VGG-Face~\cite{parkhi2015deep} in a black-box manner.
From adversarial examples generated by R-101-S, we use \texttt{fc7} as the embedding and select the images with normalized L2 distance (to the corresponding benign images) beyond the threshold defined previously.
With the benign and adversarial examples,
we first extract attribute witnesses with our aligned face images and then leverage them to build a attribute-steered model.
When misclassifying $10\%$ benign inputs into adversarial images, it only correctly identifies $8\%$ adversarial images from \StAdv and $12\%$ from CW.

Moreover, we evaluate \StAdv on existing adversarial training based defense (the detailed setting is presented in supplementary materials). 
We find that accuracy of adversarial training based defense method is 10\% against the adversarial examples generated by \StAdv, while is 46.7\% against the adversarial examples generated by PGD~\cite{madry2017towards}.
It indicates that existing adversarial training based defense method is less effective against \StAdv, which further demonstrates that our \StAdv identifies an unexplored research area beyond previous $L_p$-based ones.

\cutsubsectionup
\subsection{\StAdv on Face Landmark Detection}
\cutsubsectiondown

\begin{figure}[t]
    \centering
    \includegraphics[width=0.95\linewidth]{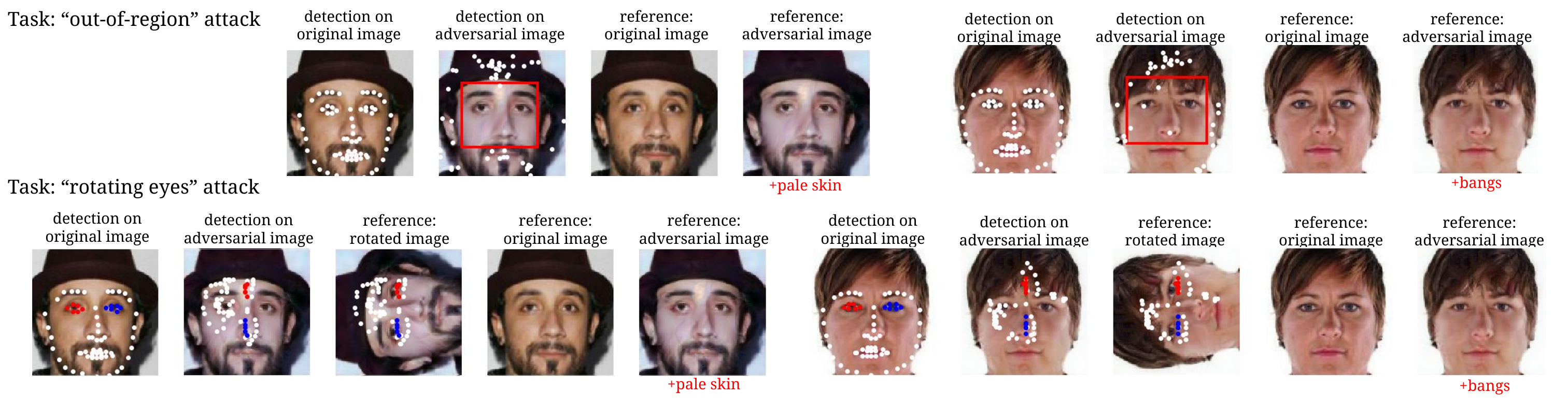}
    \cutfigurecaptionup
    \caption{Qualitative results on attacking face landmark detection model}
    \label{fig:single_landmark_detection}
    \cutfigurecaptiondown
\end{figure}

We evaluate the effectiveness of \StAdv on face landmark detection under two attack tasks, namely, ``Rotating Eyes'' and ``Out of Region''. 
For the ``Rotating Eyes'' task, we rotate the coordinates of the eyes in the image counter-clockwise by 90$\degree$.
For the ``Out of Region'' task, we set a target bounding box and attempt to push all points out of the box.
Figure~\ref{fig:single_landmark_detection} indicates that our method is applicable to attack landmark detection models. 

\cutsubsectionup
\subsection{\StAdv on Street-view Semantic Segmentation}
\cutsubsectiondown

\begin{figure}[t]
    \centering
    \includegraphics[width=0.98\linewidth]{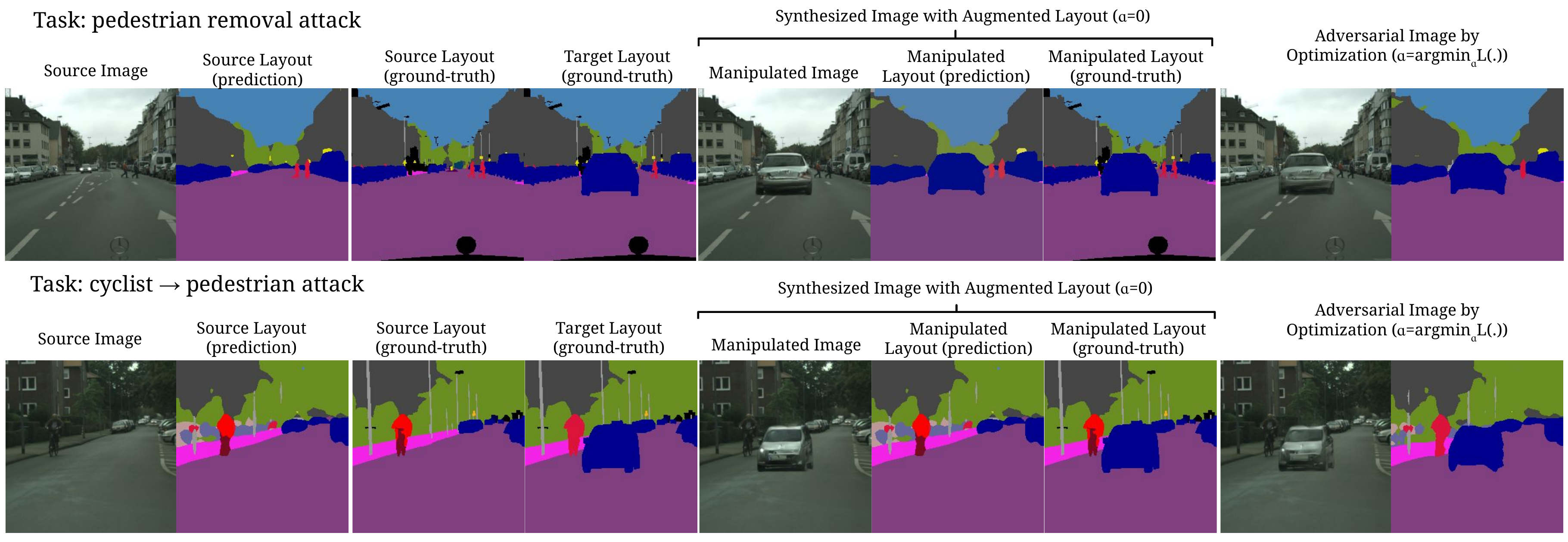}
    \cutfigurecaptionup
    \caption{Qualitative results on attacking street-view semantic segmentation model}
    \label{fig:attack_cityscale}
    \cutfigurecaptiondown
\end{figure}

We further demonstrate the applicability of our \StAdv beyond the face domain by generating adversarial perturbations on street-view images.
Figure~\ref{fig:attack_cityscale} illustrates the adversarial examples on semantic segmentation.
In the first example, we select the leftmost pedestrian as the target object instance and insert another car into the scene to attack it.
The segmentation model has been successfully attacked to neglect the pedestrian (see last column), while it does exist in the scene (see second-to-last column).
In the second example, we insert an adversarial car in the scene by \StAdv and the cyclist has been recognized as a pedestrian by the segmentation model.

\cutsectionup
\section{Conclusions}
\cutsectiondown

Overall we presented a novel attack method \StAdv, which is capable of generating semantically meaningful adversarial perturbations guided by single semantic attribute.
Compared to existing methods, \StAdv works in a more controllable fashion.
Experimental evaluations on face verification and landmark detection demonstrate several unique properties including attack transferability.
We believe this work would open up new research opportunities and challenges in the field of adversarial learning. 
For instance, how to leverage semantic information to defend against such attacks will lead to potential new discussions.

\noindent \textbf{Acknowledgement}
This work was supported in part by the National Science Foundation under Grant CNS-1422211, CNS-1616575, IIS-1617767, DARPA under Grant 00009970, and Google PhD Fellowship to X.~Yan.

\clearpage
%
%
\bibliographystyle{splncs04}
\bibliography{references}

\clearpage
\appendix

\cutsectionup
\section{Implementation details}
\cutsectiondown
In this section, we provide implementation details used in our experiments.
We implement our \StAdv using PyTorch~\cite{paszke2017automatic}.
Our implementation will be available after the final decision.

\cutsubsectionup
\subsection{Face identity verification}

We use Adam optimizer~\cite{kingma2014adam} to generate adversarial examples for both our \StAdv and the pixel-wise attack method CW~\cite{carlini2017towards}.
More specifically, we run optimization for up to $200$ steps with a fixed updating rate $0.05$ under G-FPR $< {10}^{-4}$.
Under cases with a slightly higher G-FPR, we run the optimization for up to $500$ steps with a fixed updating rate $0.01$.
For the pixel-wise attack method CW, we use additional pixel reconstruction objective with the weight set to $5$.
Specifically, we run optimization for up to $1,000$ steps with a fixed updating rate $10^{-3}$.

\cutparagraphup
\paragraph{Evaluation metrics.}
To evaluate the performance of \StAdv under different attributes,
we consider three metrics as follows:
\begin{itemize}
    \item {\em Best}: 
    the attack is successful as long as one single attribute among 17 can be successfully attacked; 
    \item {\em Average}: we calculate the average attack success rate among 17 attributes for the same face identity;
    \item {\em Worst}: 
    the attack is successful only when all of 17 attributes can be successfully attacked;
\end{itemize}
Please note that we use the {\em Best} metric as a fair comparison to the attack success rate reported by existing pixel-wise attack methods, while \StAdv can be generated with different attributes as one of our advantages.
In practice, both our \StAdv ({\em Best}) and CW achieve 100\% attack success rate.
In addition, we report the performance using the {\em average} and {\em worst} metric, which enables us to analyze the adversarial robustness towards certain semantic attributes.

\cutparagraphup
\paragraph{Pixel-wise defense methods.}
\texttt{Feature squeezing}~\cite{xu2017feature} is a simple but effective method by reducing color bit depth to remove the adversarial effects. We compress the image represented by 8 bits for each channel to 4 bits for each channel to evaluate the effectiveness.
For \texttt{Blurring}~\cite{li2017adversarial}, we use a $3 \times 3$ Gaussian kernel with standard deviation 1 to smooth the adversarial perturbations.
For \texttt{JPEG}~\cite{dziugaite2016study}, it leverages the compression and decompression to remove the adversarial perturbation. We set the compression ratio as $0.75$ in our experiment.

\subsection{Face landmark detection}
We use Adam optimizer~\cite{kingma2014adam} to generate \StAdv against the face landmark detection model.
Specifically, we run optimization for up to $2,000$ steps with a fixed updating rate $0.05$ with the balancing factor $\lambda$ set to $0.01$ (see Eq.~3 in the main paper).

\cutparagraphup
\paragraph{Evaluation Metrics.}
We apply different metrics for two adversarial attack tasks, respectively.
For ``Rotating Eyes'' task, we use a widely adopted metric \textit{Normalized Mean Error (NME)}~\cite{bulat2017far} for experimental evaluation.
\begin{equation}
    r_\text{NME} = \dfrac{1}{N} \sum_{k=1}^N \dfrac{||\mathbf{p}_k - \mathbf{\hat{p}}_k||_2}{\sqrt{W_B * H_B}},
\end{equation}
where $\mathbf{p}_k$ denotes the $k$-th ground-truth landmark, $\mathbf{\hat{p}_k}$ denotes the $k$-th  predicted landmark and $\sqrt{W_B * H_B}$ is the square-root area of ground-truth bounding box, where $W_B$ and $H_B$ represents the width and height of the box.

For ``Out of Region'' task, we consider the attack is successful if the landmark predictions fall outside a pre-defined centering region on the portrait image.
We introduce a metric that reflects the portion of landmarks outside of the pre-defined centering region: $r_\text{OR} = \frac{N_\text{out}}{N_\text{total}}$, where $N_\text{out}$ denotes the number of predicted landmarks outside the pre-defined bounding box and $N_\text{total}$ denotes the total number of landmarks.

\cutsubsectionup
\subsection{Ablation study: feature-space interpolation}
\cutsubsectiondown


%
%

\begin{figure}[ht]
    \centering
    \includegraphics[width=0.75\linewidth]{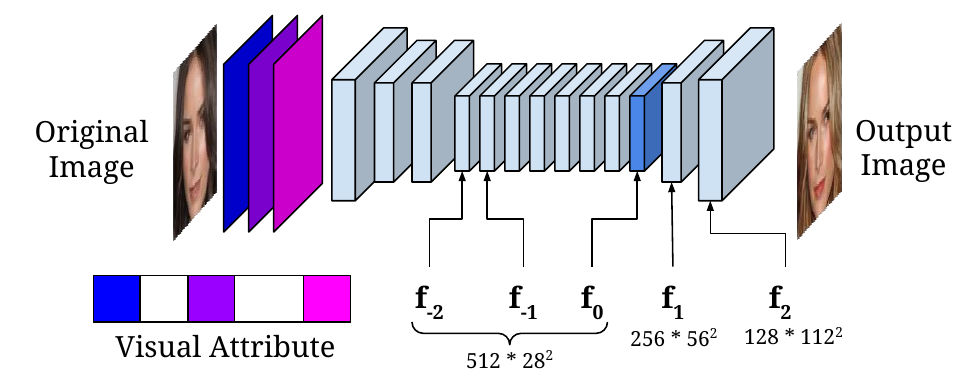}
    \cutfigurecaptionup
    \caption{The illustration of the features we used in StarGAN encoder-decoder architecture.}
    \label{fig:vis_stargan}
    \cutfigurecaptiondown
\end{figure}

We include an ablation study on feature-space interpolation by analyzing attack success rates using different feature-maps in the main paper.
We illustrate the choices of StarGAN feature-maps used in Figure~\ref{fig:vis_stargan}.
%
%
Table~1 in the main paper shows the attack success rate on R-101-S. As shown in Figure~\ref{fig:vis_stargan}, we use $\mathbf{f}_i$ to represent the feature-map after $i$-th up-sampling operation. $\mathbf{f}_0$ denotes the feature-map before applying up-sampling operation.
%
The result demonstrates that samples generated by interpolating on $\mathbf{f}_0$ achieve the highest success rate.
Since $\mathbf{f}_0$ is the feature-map before decoder, it still well embeds semantic information in the feature space.
We adopt $\mathbf{f}_0$ for interpolation in our experiments.

\cutsectionup
\section{Additional quantitative results}
\cutsectiondown

\subsection{Face identity verification}

\paragraph{Benchmark performance.}
We provide additional information about the ResNet models used in the experiments.
Table~\ref{tab:face_verify_benchmark} illustrates the performance on multiple face identity verification benchmarks including Labeled Face in the Wild (LFW) dataset~\cite{huang2008labeled}, AgeDB-30 dataset~\cite{moschoglou2017agedb}, and Celebrities in Frontal-Profile (CFP) dataset~\cite{sengupta2016frontal}.
LFW~\cite{huang2008labeled} is the \emph{de facto} standard testing set for face verification under unconstrained conditions, which contains $13,233$ face images from $5,749$ identities. 
AgeDB~\cite{moschoglou2017agedb} contains $12,240$ images from $440$ identities. AgeDB-30 is the most challenging subsets for evaluating face verification models. The large variations in age makes the face model perform worse on this dataset than on LFW.
CFP~\cite{sengupta2016frontal} consists of $500$ identities, where each identity has 10 frontal and 4 profile images. Although good performance has been achieved on the Frontal-to-Frontal (CFP-FF) test protocol, the Frontal-to-Profile (CFP-FP) test protocol still remains challenging as most of the face training sets have very few profile face images.
Table~\ref{tab:face_verify_benchmark} indicates that the used face verification model achieves state-of-the-art under all benchmarks.

\begin{table}[h]
    \cuttablecaptionup
    \caption{The verification accuracy (\%) of ResNet models on multiple face recognition datasets including LFW, AgeDB-30, and CFP.}
    \centering
    \scalebox{1.0}{\begin{tabular}{l|c|c|c|c}
    \toprule
    $\mathcal{M}$ $/$ benchmarks & LFW & AgeDB-30 & CFP-FF & CFP-FP \\
    \midrule
    R-50-S & 99.27 & 94.15 & 99.26 & 91.49 \\
    R-101-S & 99.42 & 95.93 & 99.57 & 95.07 \\
    R-50-C & 99.38 & 95.08 & 99.24 & 90.24 \\
    R-101-C & 99.67 & 95.58 & 99.57 & 92.71\\
    \bottomrule
    \end{tabular}}
    \label{tab:face_verify_benchmark}
\end{table}

\cutparagraphup
\paragraph{Thresholds for identity verification.}
To decide whether two portrait images belong to the same identity or not, we use the normalized $L_2$ distance between face features and set the FPR thresholds accordingly, which is a commonly used procedure when evaluating the face verification model~\cite{klare2015pushing,kemelmacher2016megaface}.
Table~\ref{tab:face_id_threshold} illustrates the threshold values used in our experiments when determining whether two portrait images belong to the same identity or not.
\begin{table}[th]
    \cuttablecaptionup
    \caption{
    The threshold values for face identity verification.
    }
    \centering
    \begin{tabular}{l|c|c|c|c}
    \toprule
    FPR/$\mathcal{M}$ & R-50-S & R-101-S & R-50-C & R-101-C \\
    \midrule
    $10^{-3}$ & 1.181 & 1.244 & 1.447 & 1.469 \\
    $3 \times 10^{-4}$ & 1.058 & 1.048 & 1.293 & 1.242 \\
    $10^{-4}$ & 0.657 & 0.597 & 0.864 & 0.809 \\
    \bottomrule
    \end{tabular}
    \label{tab:face_id_threshold}
\end{table}

\paragraph{Quantitative analysis.}
Combining the results from Table~\ref{tab:appendix_face_verify} and Figure~4 in the main paper, we understand that the face verification models used in our experiments have different levels of robustness across attributes. For example, face verification models are more robust against local shape variations than color variations, e.g., pale skin has higher attack success rate than mouth open. We believe these discoveries will help the community further understand the properties of face verification models.

Table~\ref{tab:appendix_face_verify} shows the overall performance (accuracy) of face verification model and attack success rate of \StAdv and CW.
As shown in Table~\ref{tab:appendix_face_verify}, although the face model trained with \texttt{cos} objective achieves higher face recognition performance, it is more vulnerable to adversarial attack compared with the model trained with \texttt{softmax} objective.
Table~\ref{tab:appendix_face_verify_2} shows that the intermediate results of \StAdv before adversarial perturbation cannot attack successfully, which indicates the success of \StAdv comes from adding adversarial perturbations through interpolation.

\begin{table}[h!]
    \cuttablecaptionup
    \caption{
    Quantitative results of identity verification (\%). It shows accuracy of face verification model and attack success rate of \StAdv and CW.
    }
    \centering
    \begin{tabular}{c|l|c|c|c|c}
    \toprule
    G-FPR & Metrics $/$ $\mathcal{M}$ & R-50-S & R-101-S & R-50-C & R-101-C \\
    \midrule
    \multirow{5}{*}{$10^{-3}$} & Verification Accuracy & 98.36 & 98.78 & 98.63 & 98.84 \\
    \cline{2-6}
    & \StAdv({\em Best}) & 100.00 & 100.00 & 100.00 & 100.00 \\
    & \StAdv({\em Worst}) & 91.95 & 93.98 & 99.53 & 99.77 \\
    & \StAdv({\em Average}) & 98.98 & 99.29 & 99.97 & 99.99 \\
    \cline{2-6}
    & CW & 100.00 & 100.00 & 100.00 & 100.00 \\
    \hline
    \multirow{5}{*}{$3 \times 10^{-4}$} & Verification Accuracy & 97.73 & 97.97 & 97.91 & 97.85 \\
    \cline{2-6}
    & \StAdv({\em Best}) & 100.00 & 100.00 & 100.00 & 100.00 \\
    & \StAdv({\em Worst}) & 83.75 & 79.06 & 98.98 & 96.64 \\
    & \StAdv({\em Average}) & 97.72 & 97.35 & 99.92 & 99.72 \\
    \cline{2-6}
    & CW & 100.00 & 100.00 & 100.00 & 100.00 \\
    \hline
    \multirow{5}{*}{$10^{-4}$} & Verification Accuracy & 93.25 & 92.80 & 93.43 & 92.98 \\
    \cline{2-6}
    & \StAdv({\em Best}) & 100.00 & 100.00 & 100.00 & 100.00 \\
    & \StAdv({\em Worst}) & 33.59 & 19.84 & 67.03 & 48.67 \\
    & \StAdv({\em Average}) & 83.53 & 76.64 & 95.57 & 91.13 \\
    \cline{2-6}
    & CW & 100.00 & 100.00 & 100.00 & 100.00 \\
    \bottomrule
    \end{tabular}
    \label{tab:appendix_face_verify}
\end{table}

\begin{table}[h!]
    \cuttablecaptionup
    \caption{
    Attack success rate of the intermediate output of \StAdv (\%). $\mathbf{x'}$, $G(\mathbf{x'},\mathbf{c})$ and $G(\mathbf{x'},\mathbf{c}^\text{new})$ are the intermediate results of our method before adversarial perturbation.}
    \centering
    \begin{tabular}{c|l|c|c|c|c}
    \toprule
    G-FPR & Metrics $/$ $\mathcal{M}$ & R-50-S & R-101-S & R-50-C & R-101-C \\
    \midrule
    \multirow{3}{*}{$10^{-3}$} & $\mathbf{x'}$ & 0.00 & 0.00 & 0.08 & 0.00 \\
    & $G(\mathbf{x'},\mathbf{c})$ & 0.00 & 0.00 & 0.00 & 0.23 \\
    & $G(\mathbf{x'},\mathbf{c}^\text{new})$({\em Best}) & 0.16 & 0.08 & 0.16 & 0.31 \\
    \hline
    \multirow{3}{*}{$3 \times 10^{-4}$} & $\mathbf{x'}$ & 0.00 & 0.00 & 0.00 & 0.00 \\
    & $G(\mathbf{x'},\mathbf{c})$ & 0.00 & 0.00 & 0.00 & 0.00 \\
    & $G(\mathbf{x'},\mathbf{c}^\text{new})$({\em Best}) & 0.00 & 0.00 & 0.00 & 0.00 \\
    \hline
    \multirow{3}{*}{$10^{-4}$} & $\mathbf{x'}$ & 0.00 & 0.00 & 0.00 & 0.00 \\
    & $G(\mathbf{x'},\mathbf{c})$ & 0.00 & 0.00 & 0.00 & 0.00 \\
    & $G(\mathbf{x'},\mathbf{c}^\text{new})$({\em Best}) & 0.00 & 0.00 & 0.00 & 0.00 \\
    \bottomrule
    \end{tabular}
    \label{tab:appendix_face_verify_2}
\end{table}

\subsection{Face landmark detection}

\begin{table}[th]
    \caption{Quantitative results on face landmark detection (\%) The two row shows the measured ratios (lower is better) for ``Rotating Eyes'' and ``Out of Region'' task, respectively.}
    \centering
    \scalebox{0.85}{
    \begin{tabular}{l|c|c|c|c|c|c|c|c|c}
        \toprule
        \multirow{2}{*}{Tasks (Metrics)}& \multirow{2}{*}{Pristine} & \multicolumn{8}{c}{Augmented Attributes }\\ \cline{3-10}
        & & \tabcell{Blond\\Hair} & Young &
                    Eyeglasses & \tabcell{Rosy\\Cheeks} &
                    Smiling & \tabcell{Arched\\Eyebrows} &
                    Bangs & \tabcell{Pale\\Skin}\\
        \midrule
        $r_\text{NME}$ $\downarrow$ & 28.04 & 14.03 & 17.28 & 8.58 & 13.24 & 19.21 & 23.42 & 15.99 & 10.72 \\
        \hline
        $r_\text{OR}$ $\downarrow$ & 45.98 & 17.42 & 23.04 & 7.51 & 16.65 & 25.44 & 33.85 & 20.03 & 13.51 \\
        \bottomrule
    \end{tabular}
    }
    \label{tab:exp_landmark}
\end{table}

We present the quantitative results of \StAdv on face landmark detection model in Table~\ref{tab:exp_landmark} including two adversarial tasks, namely, ``Rotating Eyes'' and ``Out of Region''.
We observe that our method is efficient to perform attacking on landmark detection models. 
For certain attributes such as ``Eyeglasses'' and ``Pale Skin'', \StAdv achieves reasonably-good performance.

\subsection{User study}

\begin{figure}[h!]
    \centering
    \includegraphics[width=0.85\linewidth]{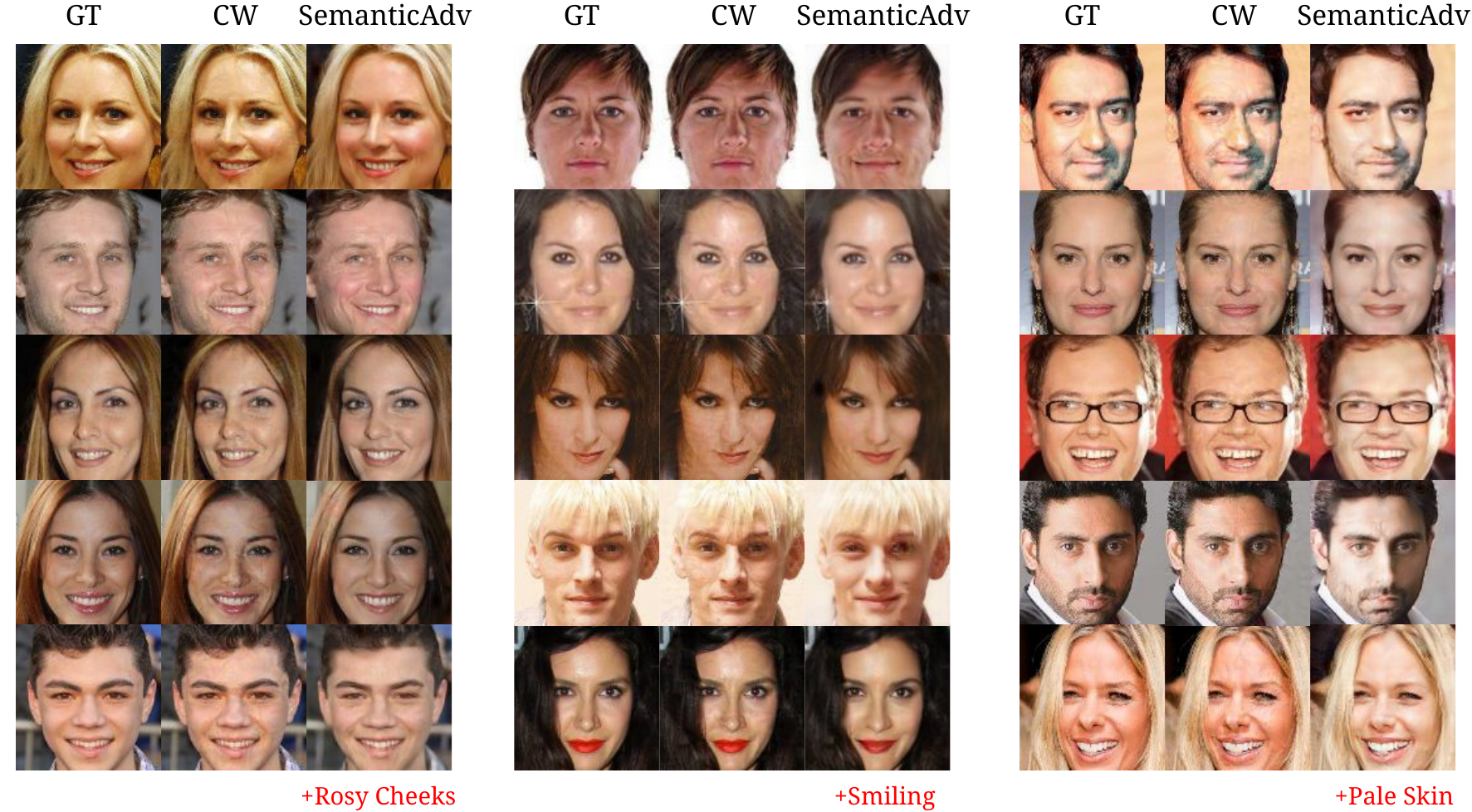}
    \cutfigurecaptionup
    \caption{Qualitative comparisons among ground truth, pixel-wise adversarial examples generated by CW, and our proposed \StAdv. Here, we present the results from G-FPR  $<10^{-4}$ so that perturbations are visible.
    }
    \label{fig:figure_appendix_user_study}
    \cutfigurecaptiondown
\end{figure}

We conduct a user study on the adversarial images of \StAdv and CW used in the experiment of API-attack and the original images.
The adversarial images are generated with G-FPR$<10^{-4}$ for both methods.
We present a pair of original image and adversarial image to participants and ask them to rank the two options.
The order of these two images is randomized and the images are displayed for 2 seconds in the screen during each trial. After the images disappear, the participants have unlimited time to select the more reasonably-looking image according to their perception.
To maintain the high quality of the collected responses, each participant can only conduct at most 50 trials, while
and each adversarial image was shown to 5 different participants.
We present the images we used for user study in Figure~\ref{fig:figure_appendix_user_study}.
In total, we collect $2,620$ annotations from 77 participants. In $39.14\pm1.96 \%$ of trials the adversarial images generated by \StAdv are selected as reasonably-looking images and in $30.27\pm1.96\%$ of trails, the adversarial images generated by CW are selected as reasonably-looking images. 
It indicates that our semantic adversarial examples are more perceptual reasonably-looking than CW.
Additionally, we also conduct the user study with larger G-FPR$=10^{-3}$. In $45.42\pm1.96 \%$ of trials, the adversarial images generated by \StAdv are selected as reasonably-looking images, which is very close to the random guess ($50\%$).

\subsection{Semantic attack transferability}
In Table~\ref{tab:trans4}, we present the quantitative results of the attack transferability under the setting with G-FPR = ${10}^{-4}$ and T-FPR = ${10}^{-4}$. 
We observe that with more strict testing criterion (lower T-FPR) of the verification model, the transferability becomes lower across different models.

\label{sup:trans}

\begin{table}[t]
    \caption{Transferability of \StAdv: cell $(i,j)$ shows attack success rate of adversarial examples generated against $j$-th model and evaluate on $i$-th model. Results are generated with G-FPR = ${10}^{-4}$ and T-FPR = ${10}^{-4}$.
    }
    \centering
    \scalebox{0.95}{\begin{tabular}{c|c|c|c|c}
    \toprule
    $\mathcal{M}_\text{test}$ $/$ $\mathcal{M}_\text{opt}$ & R-50-S & R-101-S & R-50-C & R-101-C \\
    \midrule
    R-50-S & 1.000 & 0.005 & 0.000 & 0.000 \\
    R-101-S & 0.000 & 1.000 & 0.000 & 0.000\\
    R-50-C & 0.000 & 0.000 & 1.000 & 0.000\\
    R-101-C & 0.000 & 0.000 & 0.000 & 1.000\\
    \bottomrule
    \end{tabular}}
    \label{tab:trans4}
\end{table}

\begin{table}[ht]
    \caption{Transferability of \emph{StarGAN+CW}: cell $(i,j)$ shows attack success rate of adversarial examples generated against $j$-th model and evaluate on $i$-th model. Results of \StAdv are listed in brackets.
    }
    \centering
    \begin{minipage}{0.48\textwidth}
    \begin{subtable}{1.0\textwidth}
    \centering
    \scalebox{1}{\begin{tabular}{l|c}
        \toprule
        $\mathcal{M}_\text{test}$ $/$ $\mathcal{M}_\text{opt}$ & R-101-S \\
        \midrule
        R-50-S & 0.035 (0.108) \\
        \hline
        R-101-S & 1.000 (1.000) \\
        \hline
        R-50-C & 0.145 (0.202) \\
        \hline
        R-101-C & 0.085 (0.236) \\
        \bottomrule
    \end{tabular}}
    \cuttablecaptionup
    \caption{G-FPR=$10^{-3}$, T-FPR=$10^{-3}$}
    \cuttablecaptionup
    \label{tbl:trans3_new}
    \end{subtable}
    \end{minipage}
    \begin{minipage}{0.48\textwidth}
    \begin{subtable}{1.0\textwidth}
    \centering
    \scalebox{1}{\begin{tabular}{l|c}
        \toprule
        $\mathcal{M}_\text{test}$ $/$ $\mathcal{M}_\text{opt}$ & R-101-S \\
        \midrule
        R-50-S & 0.615 (0.862) \\
        \hline
        R-101-S & 1.000 (1.000) \\
        \hline
        R-50-C & 0.570 (0.837) \\
        \hline
        R-101-C & 0.695 (0.888) \\
        \bottomrule
    \end{tabular}}
    \cuttablecaptionup
    \caption{G-FPR=$10^{-4}$, T-FPR=$10^{-3}$}
    \cuttablecaptionup
    \label{tbl:highlow_new}
    \end{subtable}
    \end{minipage}
    \label{tab:transfer_analysis_appendix_new}
\end{table}

To further showcase that our \StAdv is non-trivially different from pixel-wise attack added on top of semantic image editing, we provide one additional baseline called StarGAN+CW and evaluate its attack transferability.
This baseline first performs semantic image editing using the StarGAN model (non-adversarial) and then conducts the standard $L_p$ CW attacks on the generated images.
As shown in Table~\ref{tab:transfer_analysis_appendix_new}, the StarGAN+CW baseline has noticeable performance gap to our proposed \StAdv. 
This also justifies that our \StAdv is able to produce novel adversarial examples which cannot be simply achieved by combining attribute-conditioned image editing model with $L_p$ bounded perturbation.

\cutsubsectionup
\subsection{Query-free black-box API attack}
\cutsubsectiondown

\begin{table}[t]
    \caption{Quantitative analysis on query-free black-box attack. 
    We use ResNet-101 optimized with \texttt{softmax} loss for evaluation and report the attack success rate(\%).
    Note that for Micresoft Azure API, it does not provide the accept thresholds for different T-FPRs and thus we use the provided likelihood 0.5 to determine whether two faces belong to the same person.
    }
    \centering\small
    \scalebox{1}{
    \begin{tabular}{l|c|c|c|c|c}
        \toprule
        API name & \multicolumn{2}{c|}{Face++} & \multicolumn{2}{c|}{AliYun} & \multicolumn{1}{c}{Azure} \\ \cline{2-6}
        Metric & \multicolumn{2}{c|}{T-FPR} & \multicolumn{2}{c|}{T-FPR} & \multicolumn{1}{c}{Likelihood} \\ \cline{2-6}
        Attacker $/$ Metric value & $10^{-3}$ & $10^{-4}$ & $10^{-3}$ & $10^{-4}$ & 0.5\\
        \midrule
        Original $\mathbf{x}$ & 2.04 & 0.51 & 0.50 & 0.00 & 0.00\\
        Generated $\mathbf{x^\text{new}}$ & 4.21 & 0.53& 0.50 & 0.00 & 0.00\\
        \hline
        CW (G-FPR = $10^{-3}$) & 9.18 & 2.04 & 2.00 & 0.50 & 0.00\\
        StarGAN+CW (G-FPR = $10^{-3}$) & 15.9 & 3.08 & 3.50 & \textbf{1.00} & 0.00 \\
        \StAdv (G-FPR = $10^{-3}$) & \textbf{20.00} & \textbf{4.10} & \textbf{4.00} & 0.50 & 0.00 \\
        \hline
        CW (G-FPR = $10^{-4}$) & 28.57 & 10.17 & 10.50 & 2.50 & 1.04\\
        StarGAN+CW (G-FPR = $10^{-4}$) & 35.38 & 14.36 & 12.50 & 3.50 & 1.05\\
        \StAdv (G-FPR = $10^{-4}$) & \textbf{58.25} & \textbf{31.44} & \textbf{24.00} & \textbf{10.50} & \textbf{5.73} \\
        \hline
        CW & 37.24 & 20.41 & 18.00 & 9.50 & 3.09\\
        StarGAN+CW & 47.45 & 26.02 & 20.00 & 8.50 &5.56\\
        MI-FGSM~\cite{dong2018boosting} & 53.89 & 30.57 & 29.50 & 17.50 & 10.82 \\
        M-DI$^2$-FGSM~\cite{xie2019improving} & 56.12 & 33.67 & 30.00 & 18.00 & 12.04 \\
        \StAdv (G-FPR $<10^{-4}$) & \textbf{67.69} & \textbf{48.21} & \textbf{36.5} & \textbf{19.5} & \textbf{15.63} \\
        \bottomrule
    \end{tabular}}
    \label{tab:exp_blackbox_analysis}
\end{table}

\begin{figure}[h]
    \centering
    \includegraphics[width=1.0\linewidth]{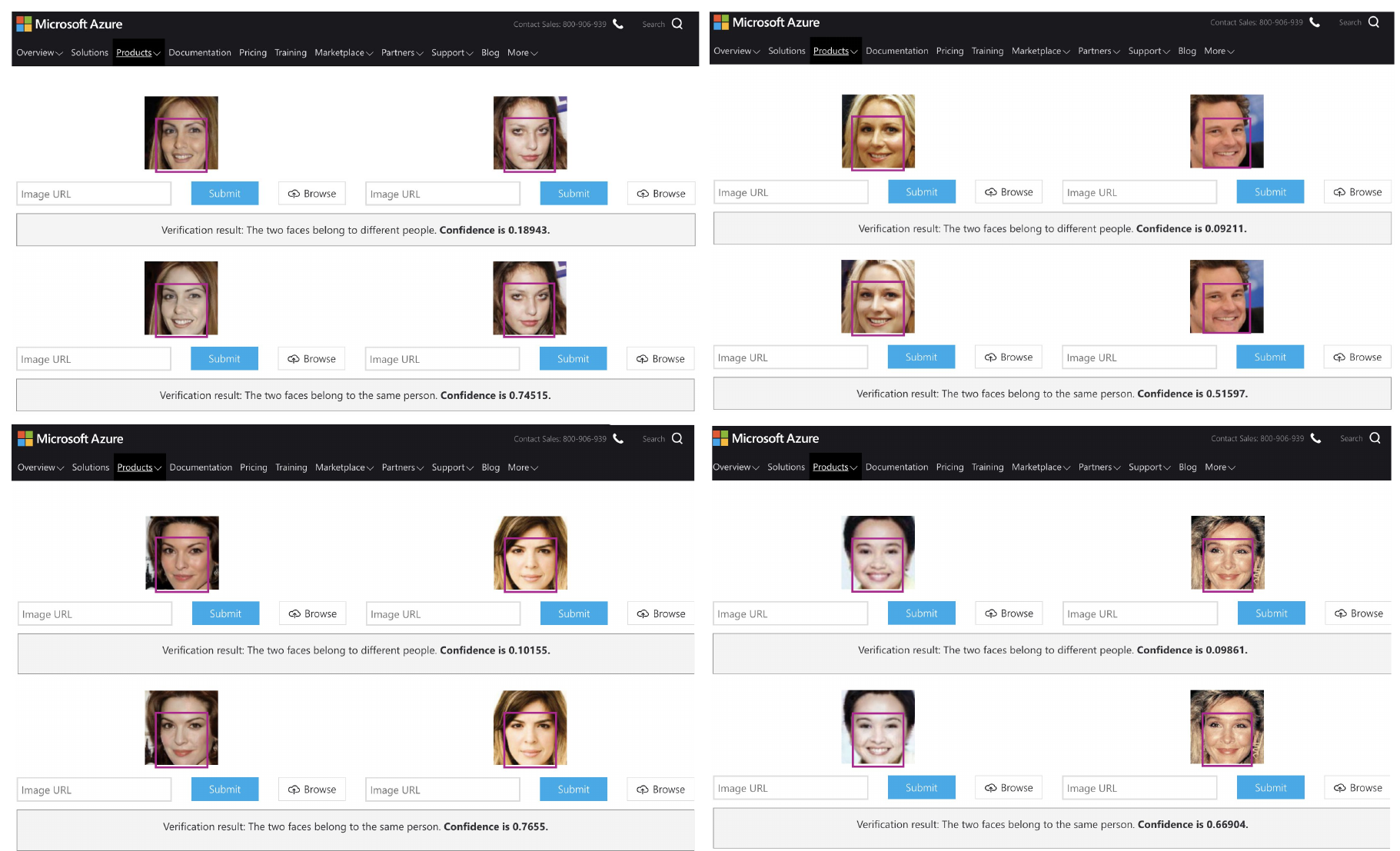}
    \cutfigurecaptionup
    \caption{Illustration of our \StAdv in the real world face verification platform (editing on pale skin). Note that the confidence denotes the likelihood that two faces belong to the same person.
    }
    \label{fig:figure_appendix_api_attack}
    \cutfigurecaptiondown
\end{figure}

In Table~\ref{tab:exp_blackbox_analysis}, we present the results of \StAdv performing query-free black-box attack on three online face verification platforms. \StAdv outperforms CW and StarGAN+CW in all APIs under all FPR thresholds. 
In addition, under the same T-FPR, we achieve higher attack success rate on APIs using samples generated using lower G-FPR compared to samples generated using higher G-FPR.
Original $\mathbf{x}$ and generated $\mathbf{x^\text{new}}$ are regarded as reference point of the performance of online face verification platforms. 
In Figure~\ref{fig:figure_appendix_api_attack}, we also show several examples of our API attack on Microsoft Azure face verification system, which further demonstrates the effectiveness of our approach.

\subsection{\StAdv against adversarial training}

We evaluate our \StAdv against the existing adversarial training based defense method~\cite{madry2017towards}.
%
In detail, we randomly sample 10 persons from CelebA~\cite{liu2015deep} and then randomly split the sampled dataset into training set, validation set and testing set according to a proportion of 80\%, 10\% and 10\%, respectively.
We train a ResNet-50~\cite{he2016deep} to identify these face images by following the standard face recognition training pipeline~\cite{sun2014deep}.
As CelebA~\cite{liu2015deep} does not contain enough images for each person, we finetune our model from a pretrained  model trained on MS-Celeb-1M~\cite{guo2016ms,zhang2018accelerated}.
We train the robust model by using  adversarial training based method ~\cite{madry2017towards}. In detail, 
we follow the same setting in ~\cite{madry2017towards}. We use 7-step PGD $L_{\infty}$ attack to generate adversarial examples to solve the inner maximum problem for adversarial training. During test process, we evaluate by using adversarial examples generated by 20-step PGD attacks. The perturbation is bounded by 8 pixel (ranging from [0,255]) in terms of  $L_{\infty}$  distance).


\begin{table}[t]
    \centering
    \cuttablecaptionup
    \caption{Accuracy on standard model (without adversarial training) and robust model (with adversarial  training). 
    }
    \scalebox{0.85}{
    \begin{tabular}{l|c|c|c}
    \toprule
    Training Method / Attack & Benign & PGD & \StAdv \\  
    \midrule 
    Standard &  93.3\% &0\% & 0\%  \\ \hline
    Robust~\cite{madry2017towards} & 86.7\% &46.7\% & 10\%\\ 
    \bottomrule
    \end{tabular}
    }
    \label{tab:advtrain}
    \cuttablecaptionup
\end{table}

As shown in Table~\ref{tab:advtrain}, the  robust model achieves  10\% accuracy against the adversarial examples generated by \StAdv, while  46.7\% against the adversarial examples generated by PGD~\cite{madry2017towards}.
It indicates that existing adversarial training based defense method is less effective against \StAdv. It further demonstrates that our \StAdv identifies an unexplored research area beyond previous $L_p$-based ones.


\clearpage
\cutsectionup
\section{Additional visualizations}
\cutsectiondown

\begin{figure}[th]
    \centering
    \includegraphics[width=0.95\linewidth]{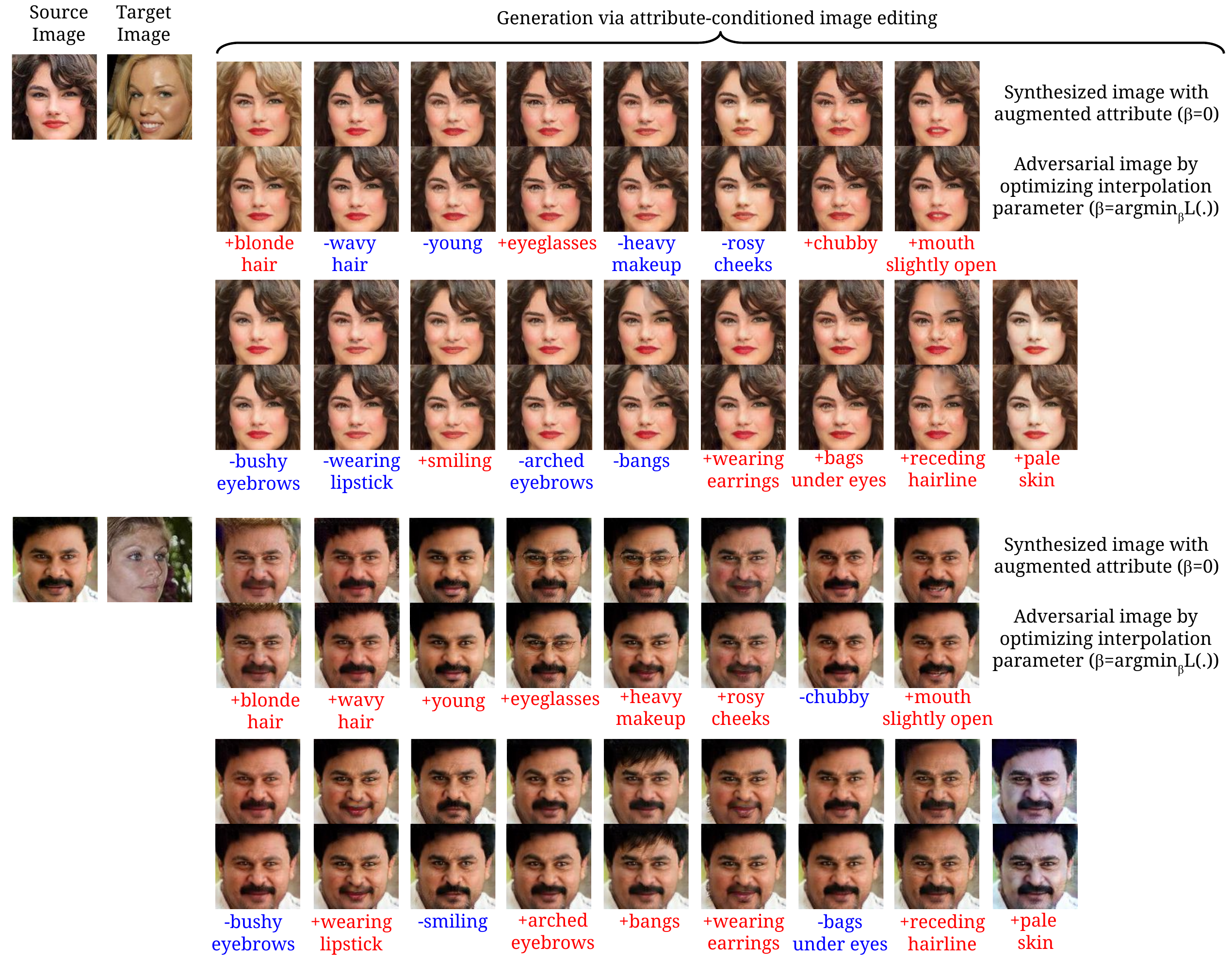}
    \includegraphics[width=0.95\linewidth]{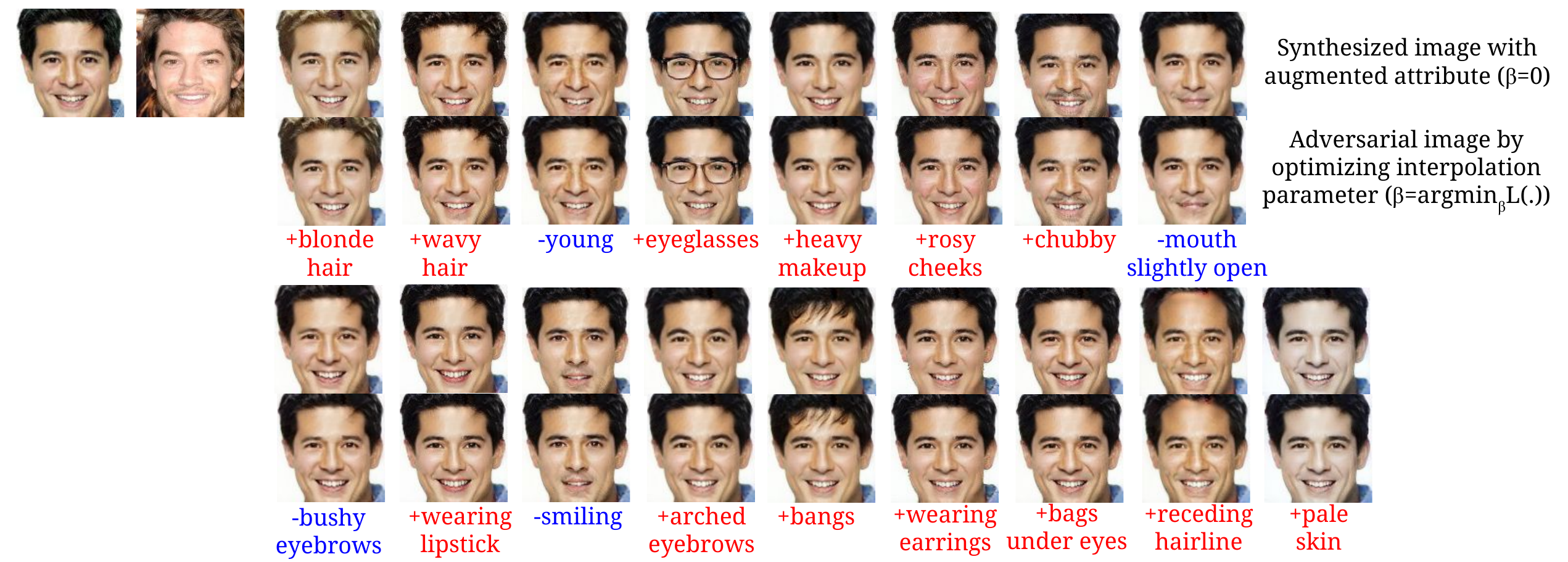}
    \caption{Qualitative analysis on single-attribute adversarial attack (G-FPR=$10^{-3}$).}
    \label{fig:vis_sing34}
\end{figure}

\begin{figure}[bh!]
    \centering
    \includegraphics[width=0.95\linewidth]{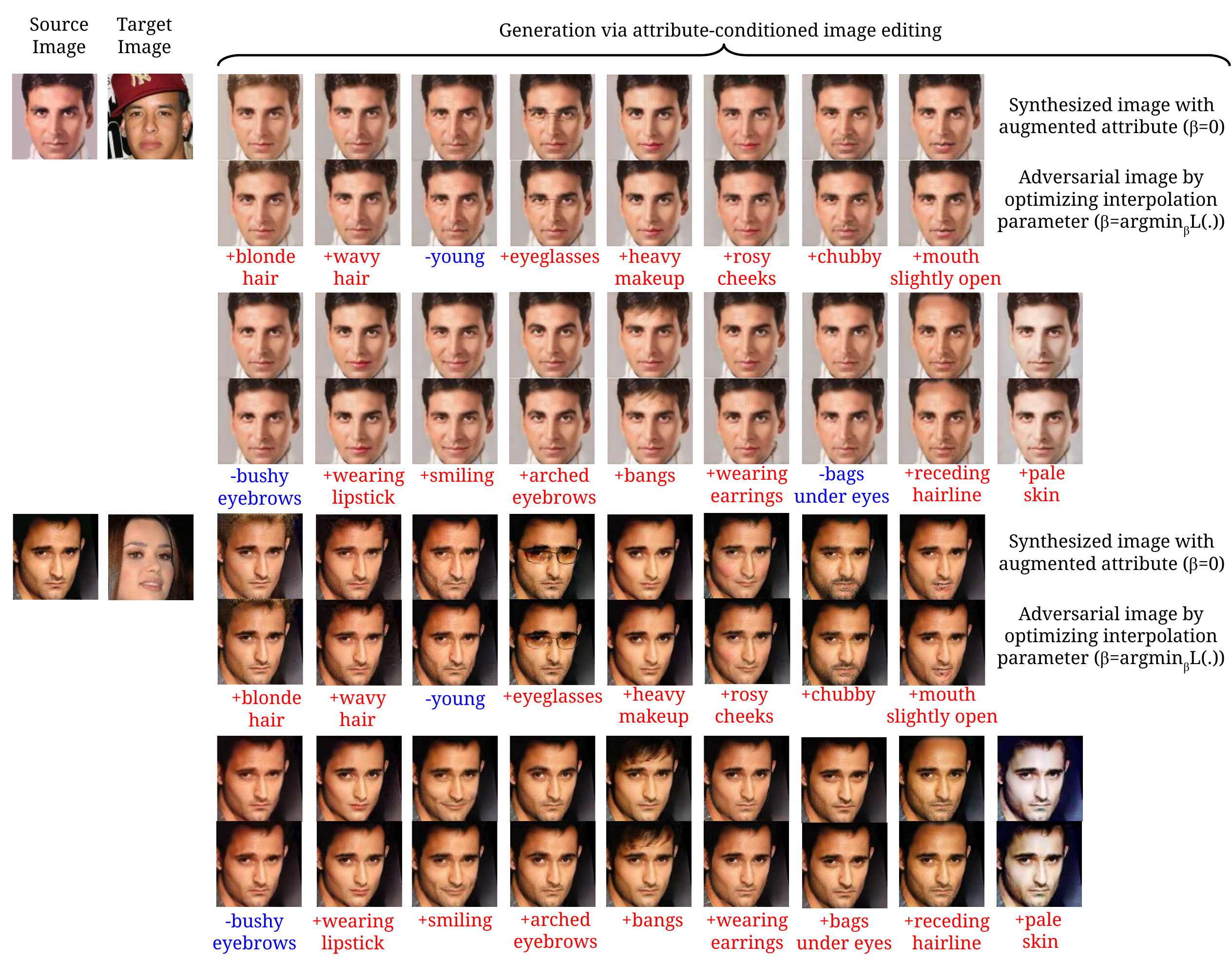}
    \includegraphics[width=0.95\linewidth]{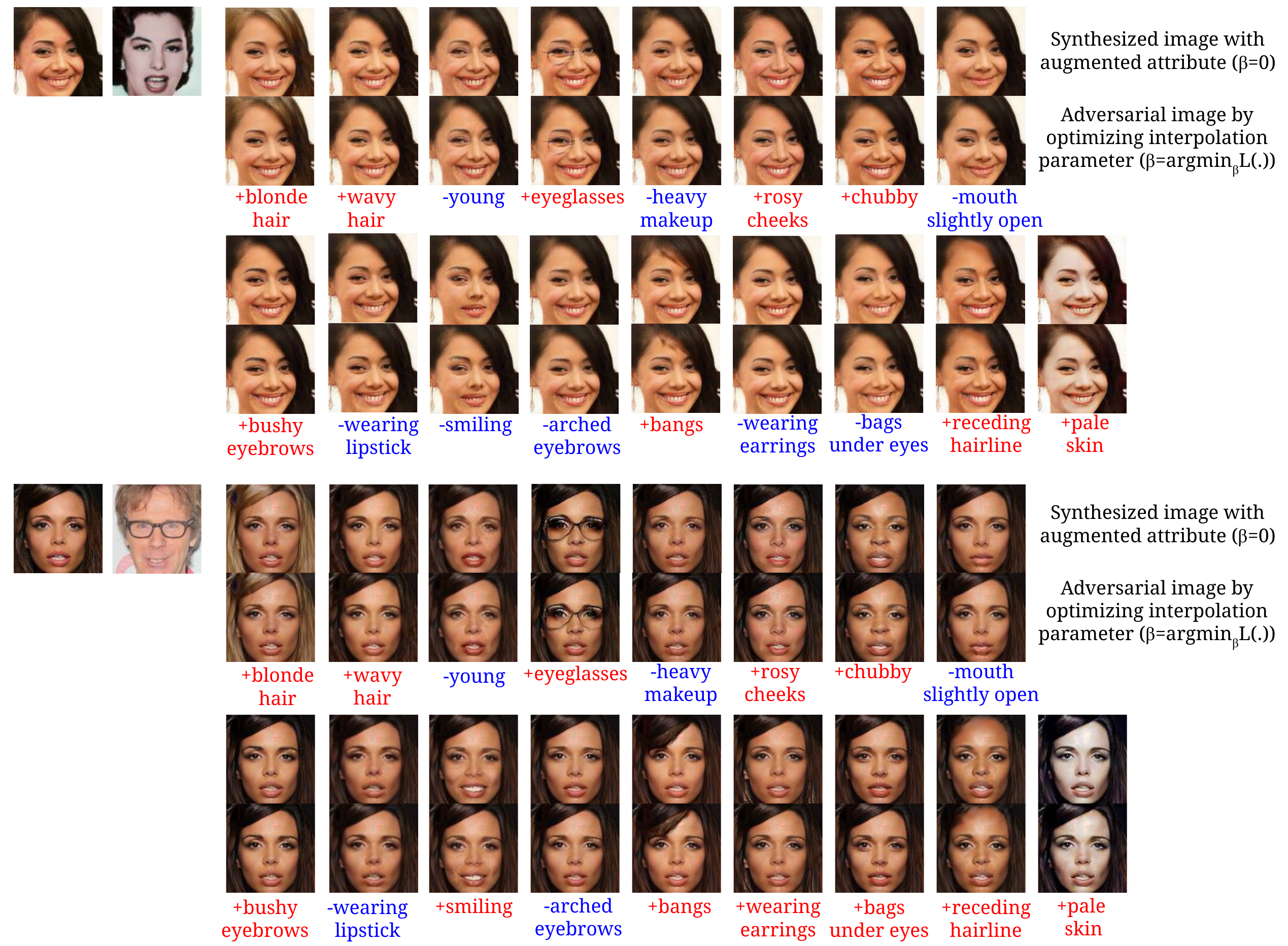}
    \caption{Qualitative analysis on single-attribute adversarial attack (G-FPR=$10^{-3}$).}
    \label{fig:vis_sing5}
    \cutfigurecaptiondown
\end{figure}

\begin{figure}[th]
    \centering
    \includegraphics[width=0.95\linewidth]{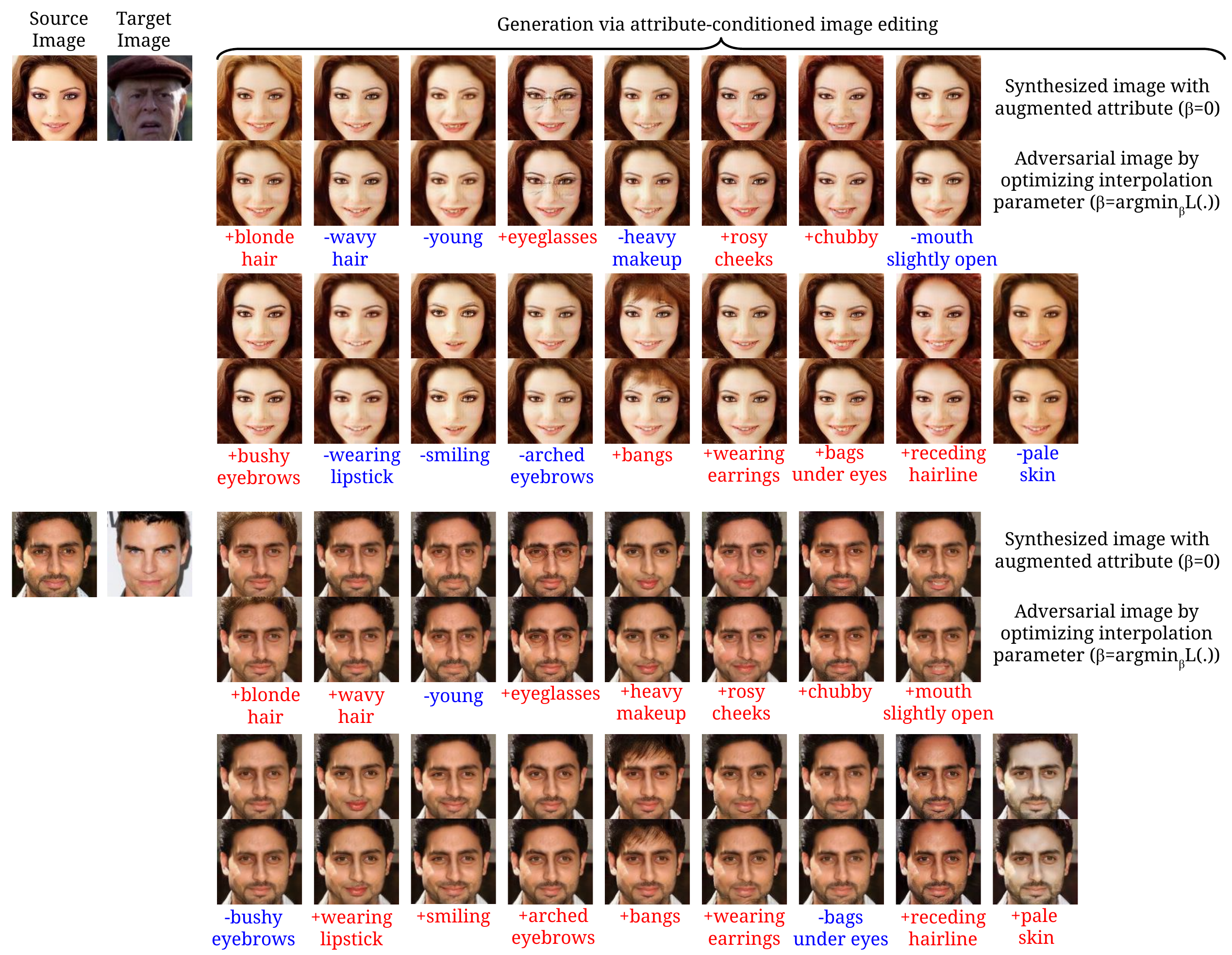}
    \includegraphics[width=0.95\linewidth]{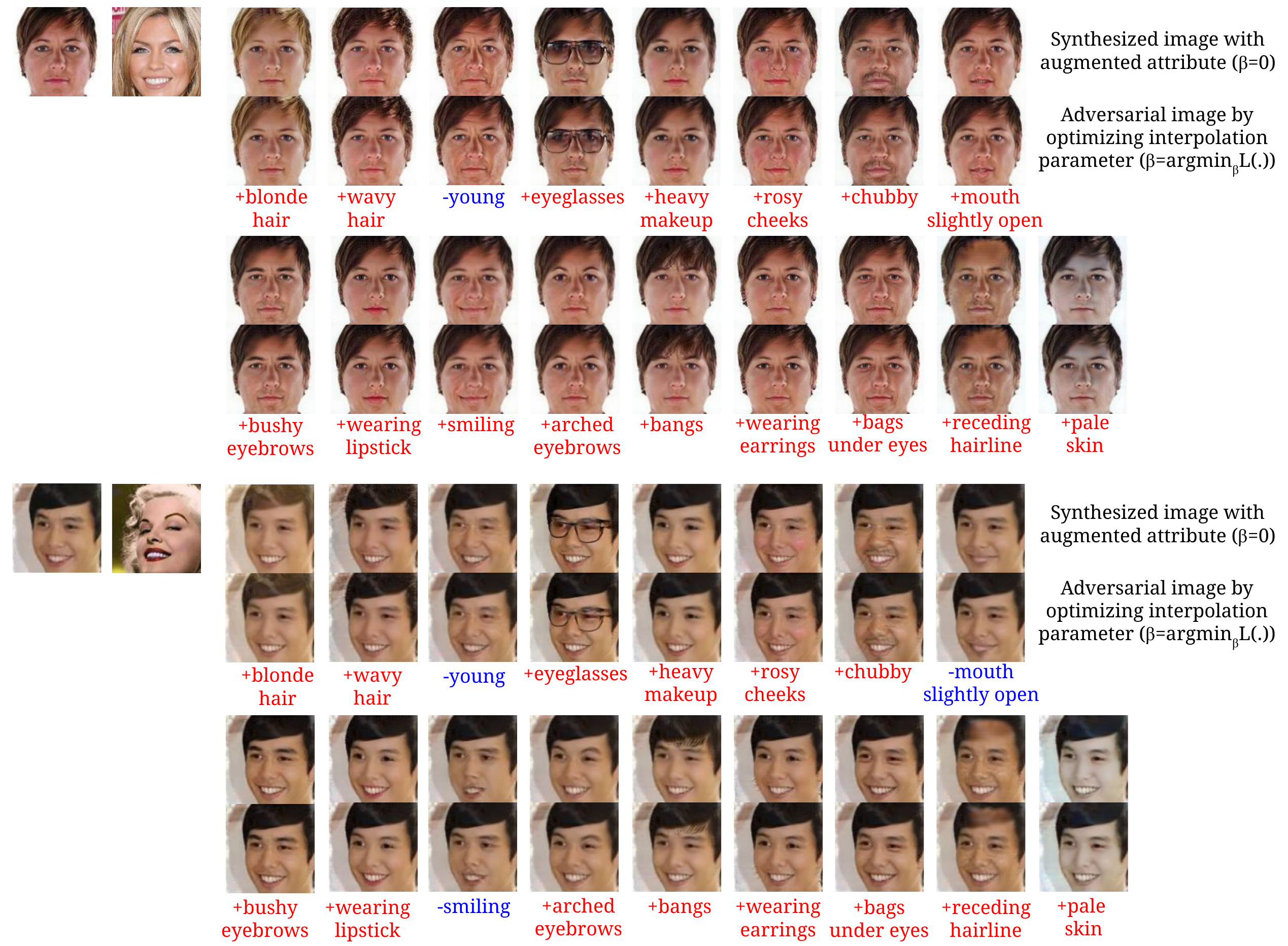}
    \caption{Qualitative analysis on single-attribute adversarial attack (G-FPR=$10^{-3}$).}
    \label{fig:vis_sing12}
\end{figure}


\begin{figure}[th]
    \centering
    \includegraphics[width=1.0\linewidth]{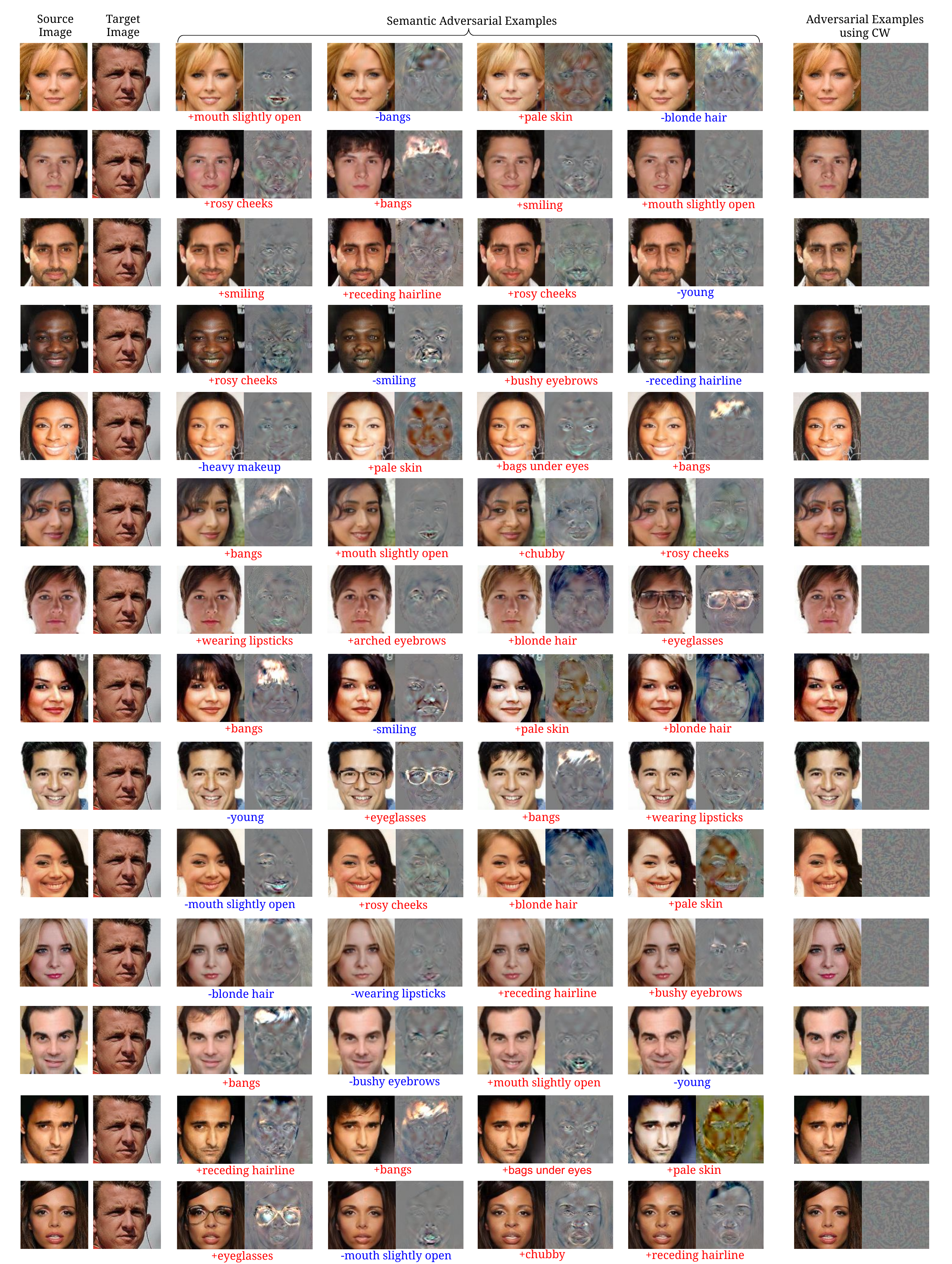}
    \caption{Qualitative comparisons between our proposed \StAdv (G-FPR = $10^{-3}$) and pixel-wise adversarial examples generated by CW. Along with the adversarial examples, we also provide the corresponding perturbations (residual) on the right.}
    \label{fig:vis_compare}
\end{figure}

\begin{figure}[th]
    \centering
    \includegraphics[width=1.0\linewidth]{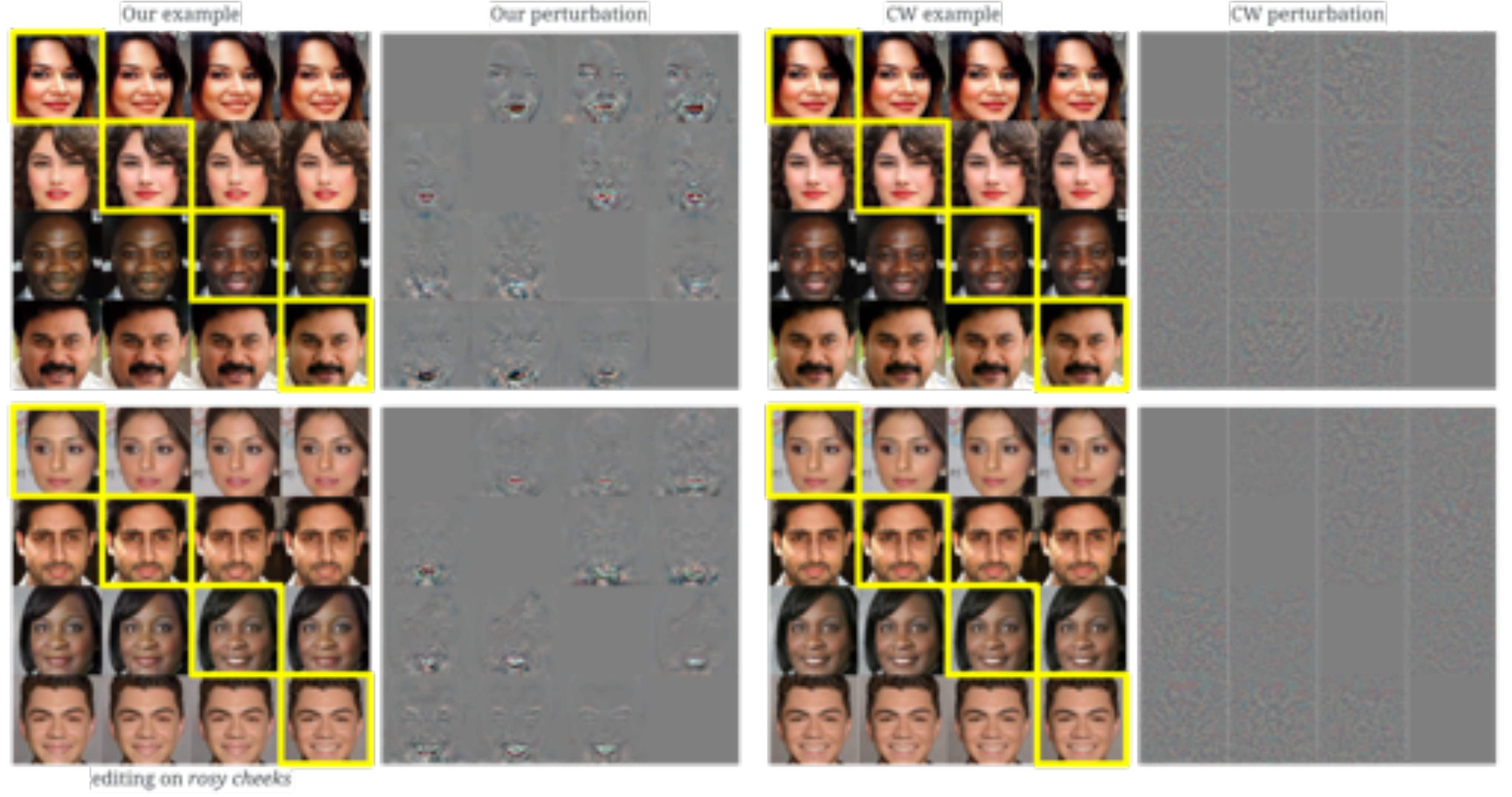}
    \includegraphics[width=1.0\linewidth]{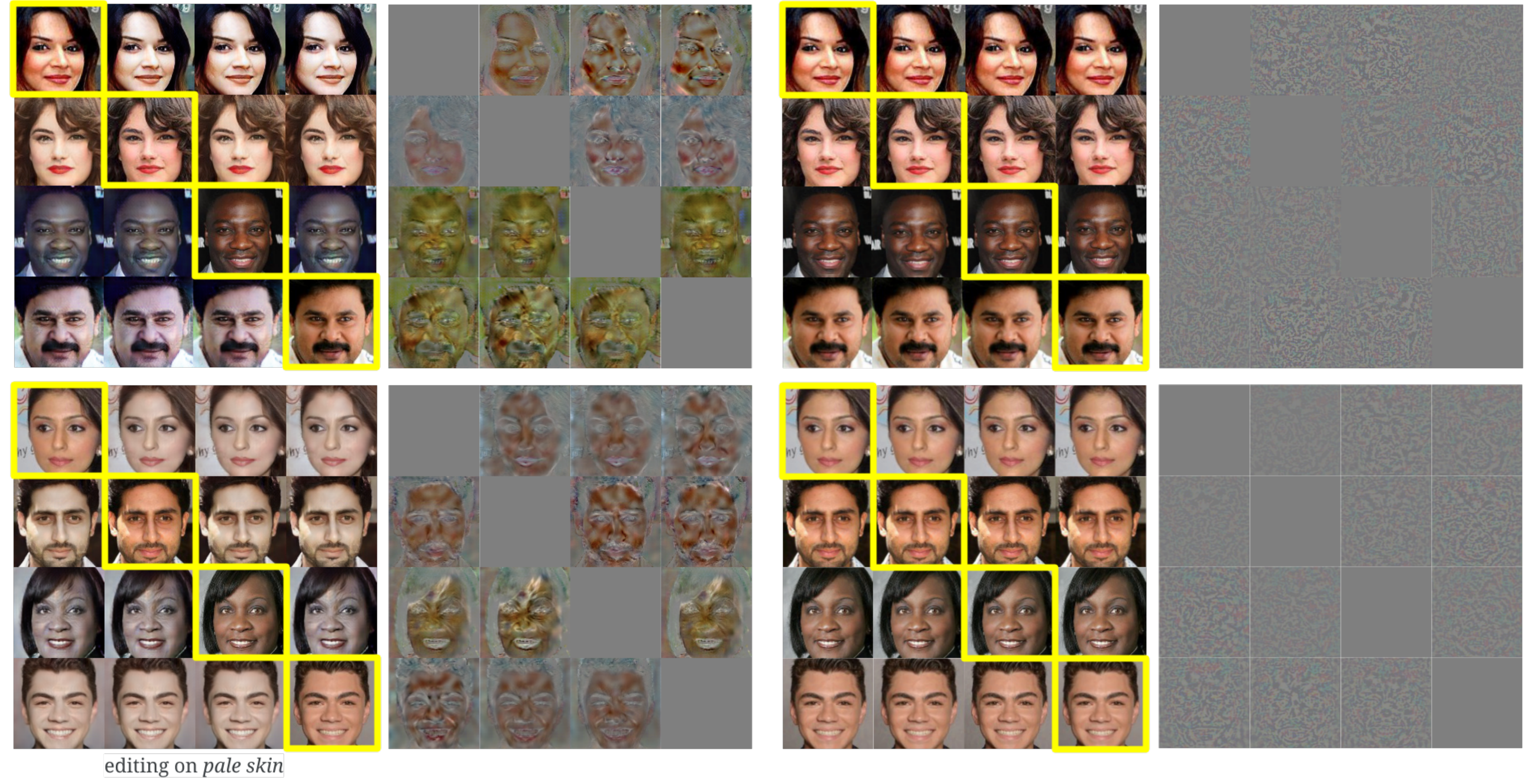}
    \caption{Qualitative analysis on single-attribute adversarial attack (\StAdv with G-FPR = $10^{-3}$) by each other. Along with the adversarial examples, we also provide the corresponding perturbations (residual) on the right.}
    \label{fig:vis_diff}
\end{figure}

\end{document}